\documentclass[preprint,12pt]{elsarticle}

\usepackage{amssymb}
\usepackage{amsmath}
\usepackage{subfig}
\usepackage{multirow}
\usepackage{lscape}
\usepackage{url}
\usepackage{graphicx}
\usepackage{epstopdf}
\biboptions{sort&compress}

\journal{Journal Name}

\begin{document}

\begin{frontmatter}



\title{
		\textbf{\LARGE Skin Diseases Detection using LBP and WLD- An Ensembling Approach}
	}


\author[1]{Arnab Banerjee}
\author[2]{Nibaran Das, Mita Nasipuri}

\address[1]{Department of Computer Science and Technology, Dr.B.C.Roy Polytechnic, Durgapur, West Bengal- 713206, India}
\address[2]{Department of Computer Science and Engineering, Jadavpur University, Kolkata, West Bengal-700032, India}

\begin{abstract}
\textbf{In all developing and developed countries in the world, skin diseases are becoming a very frequent health problem for the humans of all age groups. Skin problems affect mental health, develop addiction to alcohol and drugs and sometimes causes social isolation. Considering the importance, we propose an automatic technique to detect three popular skin diseases- Leprosy, Tinea versicolor and Vitiligo from the images of skin lesions. The proposed technique involves Weber local descriptor and Local binary pattern to represent texture pattern of the affected skin regions. This ensemble technique achieved 91.38\% accuracy using multi-level support vector machine classifier, where features are extracted from different regions that are based on center of gravity. We have also applied some popular deep learning networks such as MobileNet, ResNet\textunderscore152, GoogLeNet, DenseNet\textunderscore121, and  ResNet\textunderscore101. We get 89\% accuracy using ResNet\textunderscore101. The ensemble approach clearly outperform all of the used deep learning networks. This imaging tool will be useful for early skin disease screening.}
\end{abstract}

\end{frontmatter}

\section{Introduction}
\label{intro}
Skin Diseases has become a big concern for the people living in smart cities and industrial areas all over the world. The people with skin disease have a feeling of being cursed and try to stay away from the community. The skin diseases have a very negative impact on an individual's personal, social and working lives and also the lives of his/ her family members. The affected people suffer from depression, stress, anxiety, low self-confidence, which may culminate into a suicidal tendency. In many of the cases it has been observed that early detection of the diseases may be helpful for the doctors to fully cure a patient. In a highly populated country like India, there is a dearth of dermatologists in rural and semi-urban areas. So, keeping all these in mind an automatic, non-invasive technique to detect the skin diseases is very essential. So, keeping all these in mind an automatic, non-invasive technique to detect the skin diseases is very essential. In recent times several research works have been done to automatically detect different skin diseases from the images of skin lesions using different pattern recognition and computer vision approaches. Preparing a standard automatic or semiautomatic non-invasive technique for identifying different skin diseases is very challenging due to several reasons. Very high similarity has been observed between the lesion colors and shapes of different diseases. Also, there are significant variations in the lesion colors and shapes from person to person suffering from same disease. Another problem the researchers are facing is to the absence of a standard database of images of skin lesions for different diseases. Some attempts by the researchers to identify diseases from the images of the affected skin area using texture features are reported in \cite{article3,6686372,10.1007/978-3-642-45062-4_48,wei_gan_ji_2018}. Artificial Neural Network (ANN) and Deep Learning based approaches are used for this purpose in \cite{5409725,6498195,7026918,8030303,8233570,article,article1}. Some of these approaches use patient inputs \cite{5409725,6359626,6498195,7026918} beside skin lesion images for diagnosing the disease. The only disadvantage is that these methods requires the strong involvement of dermatologists to acquire the proper patient inputs during the treatment. In this work, we propose a ensemble approach of two popular texture features, namely, Local Binary Pattern (LBP) and Weber Local Descriptor (WLD) to enhance the performance of our previous methods\cite{6686372,10.1007/978-3-642-45062-4_48}. Local binary patterns are used to compute the features from the skin lesion images in \cite{6686372}. In \cite{10.1007/978-3-642-45062-4_48} the skin lesion images are divided into four regions using the centre of gravity and from each of four images the rotation invariant WLD features are extracted and then combined. SVM classifier is used in both \cite{6686372,10.1007/978-3-642-45062-4_48}.  In this work we have considered the images of three common skin diseases, namely, Leprosy, Vitiligo and Tinea versicolor and also the normal skin images for the experiment. We have applied grid searching for optimal parameter selection of Multi-SVM and used it for classification of skin diseases. In this work, we have compared the proposed method with Gray-Level Co-Occurrence Matrix (GLCM), Local Binary Pattern (LBP), Discrete Cosine Transform (DCT), Discrete Fourier Transform (DFT) and WLD based methods for the better establishment of the proposed approach. We have also compared the proposed method with some popular deep neural networks such as GoogLeNet, ResNet\textunderscore152, MobileNet, DenseNet\textunderscore121, ResNet\textunderscore101.
\section{Related Articles}
\label{sec2}
The existing automatic skin diseases identification techniques mainly focus on Psoriasis\cite{DelgadoGomez:2007:ACD:1238135.1238180}, Ringworm \cite{DBLP:journals/corr/abs-1103-0120} and  erythemato-squamous\cite{UBEYLI2005421}. In 2009, research-ers in \cite{5409725} designed and trained an Artificial Neural Network for skin diseases detection in a specific tropical area such as Nigeria. In this work they used patient information and causative organisms. An application of texture feature GLCM for the detection of skin diseases was proposed in \cite{article3}. They used images of three diseases classes from DERMNET imagecollection called Atopic Dermatitis, Eczema, and Urticaria. An automatic diagnostic system for six dermatological diseases (Acne, Eczema, Psoriasis, Tinea Corporis, Scabies and Vitiligo) using color skin images and different patient information such as diseased body portion, elevation and number of occurrence (in months) etc. has been proposed in \cite{6359626}. This methodology requires the involvement of dermatologist. Another dermatological diagnostic system introduced in 2012 by \cite{6498195} which was based on two feature selection method- Correlation Feature Selection (CFS), Fast Correlation-based Filter (FCBF) and BPNN (Back Propagation Based Neural Network). This method was tested on six diseases such as Psoriasis, Seboreic dermatitis, Lichen Planus, Pityriasis rosea, Cronic dermatitis and Pityriasis Rubra Pilaris. This method requires several histopathological inputs of the patient to work on. In 2013, a non-invasive automatic skin diseases detection technique using LBP as texture feature has been proposed in \cite{6686372}. They used the dataset with three skin diseases Leprosy: Tinea versicolor and Vitiligo for the experiment. Normal skin images are also considered to establish the differences among normal skin and the diseased skin. Another development was introduced by \cite{10.1007/978-3-642-45062-4_48} in 2013, for detecting Leprosy, Tinea versicolor and Vitiligo skin diseases using rotation invariant Weber local descriptor. In 2014, another method for detecting skin diseases from skin image samples using ANN has been proposed in \cite{7026918}. Different patient inputs like gender, age, liquid type and color, elevation and feeling have been taken for the identification of disease. Like \cite{6359626} this method also requires involvement of dermatologist. In 2017, a combination of Deep learning and machine learning has been employed for recognizing melanoma from dermoscopy images \cite{8030303}. An ultra-imaging technique based on photo-acoustic dermoscopy was presented by \cite{7865979} in 2017. This method uses excitation energy and penetration depth measures for skin analysis. In \cite{article2} we find review of Melanoma detection and skin monitoring techniques using smart phone apps. They also compared the performance of several apps in terms of performance and image processing techniques used. In 2017, another deep learning-based melanoma detection has been proposed in \cite{8233570}. They addressed two problem, segmentation of skin tumor from the skin and classification of skin tumor. In 2018, a non-invasive technique using GLCM for detecting three skin diseases Herpes, Dermatitis and Psoriasis has been proposed \cite{wei_gan_ji_2018}. Another work on classification of skin lesions using Deep Neural Network can be found in \cite{article}. This method only takes image pixels and corresponding disease labels as input. Recently, in 2018, a deep learning-based Melanoma detection technique has been proposed in \cite{article1}.  They have used two fully convolutional residual networks (FCRN) for lesion segmentation and lesion classification. 

\section{Proposed Approach}
\label{sec3}
\begin{figure}[ht!]
	\includegraphics[width=\linewidth]{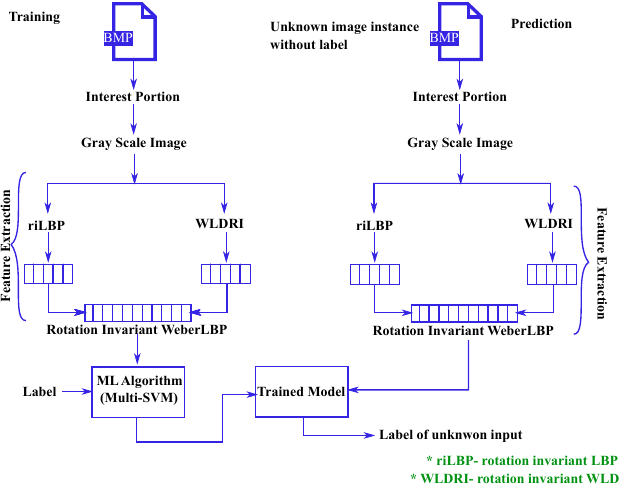}
	\caption{Framework of the proposed approach}
	\label{fig:1}
\end{figure}

The proposed approach is divided into four phases: 1. Preparation of dataset of skin lesion images for Leprosy, Vitiligo, Tinea versicolor, and normal skin images. 2. Pre-processing of the images. 3. Feature extraction from the images and 4. Classification of the skin images on the basis of extracted features using Multi-SVM with optimized parameters. The total process is presented in the figure~\ref{fig:1}.
\subsection{Dataset Preparation}
\label{subsec1}
Preparation of skin lesion image dataset for the above mentioned three skin diseases and also for the normal skin images is one the challenging part for this work. It is difficult to get access to these patients and also to obtain their consent for showing their affected skin region for taking an image. In this work, we have taken the images of skin lesions from 141 patients. We have collected the images from the outdoor patients of School of Tropical Medicine (STM), Kolkata over a period of one year starting from April, 2011 to March, 2012. In this one-year duration, we have taken the images of Leprosy, Vitiligo, and Tinea versicolor affected skin from different patients. In the case of patients having multiple skin lesions, we have collected multiple images from the patient. We have also collected the images of normal skin regions from them to differentiate between normal and diseased skins. We have deleted the blurred images from the collection and taken only the good images for the experiment. A total of 876 images are taken out of which 262 are Leprosy affected images, 210 are Vitiligo affected images, 242 are Tinea versicolr affected images and 162 are normal images. Next, we have divided the collection of images into the training and testing dataset in 4:1 proportion. We have named the dataset as CMATER Skin Dataset.   
\subsection{Pre-processing}
\label{subsec2}
The images of the dataset are cropped and resized to $144 \times 144$ to get equal proportion for the skin lesion. We convert the color images to gray scale images to make the computation easier and less time consuming. The gray scale images are used as input in the next phase. The BMP file format is used for the images in the dataset.  
\subsection{Feature Extraction}
\label{subsec3}
\subsubsection{Gray-Level Co-Occurrence Matrix (GLCM)}
\label{subsubsec1}
Texture pattern is one of the important features to distinguish different objects. Gray-Level Co-occurrence Matrix is extensively used as a texture descriptor of an image. The basic idea of Gray-Level Co-occurrence Matrix is calculating the probability of how often a pixel with an intensity value say x having a specific spatial relationship with another pixel with an intensity value say y. Basically the (x,y) entry in Gray-Level Co-occurrence matrix contains the spatial relationship probability between x and y at some distance d in the direction $\theta$. The total number of gray levels determines the size of the Gray-Level Co-occurrence matrix. The scaling technique can be used to reduce the number of gray levels in the image. In \cite{4309314} and \cite{4045583} we find the techniques for computing different statistics from the normalized symmetrical images. In the case of directional GLCM, separate co-occurrence matrices are formed for four directions- vertical ($\theta=90^0$), horizontal ($\theta=0^0$), left diagonal ($\theta=135^0$) and right diagonal ($\theta=45^0$). Now from each co-occurrence matrix, sixteen statistical measures called Energy, Entropy, Inertia, Inverse Difference Moment, Sum Average, Sum of Square Variance, Sum Entropy, Difference Average, Difference Variance, Difference Entropy, Contrast, Correlation, Information Measure of Correlation 1, Information Measure of Correlation 2, Cluster Prominence, Cluster Shade are computed. These measures are used to represent the skin lesion texture.  
\subsubsection{Discrete Cosine Transform (DCT)}
\label{subsubsec2}
The Discrete Cosine Transform (DCT) separate the image into spectral sub-bands with respect to the visual quality of the image. It is a powerful tool to extract the features from the skin lesion images. At first, the DCT is applied into the total skin lesion image and then some of the co-coefficients are used for making the feature vector. For a $M \times N$ dimensional image, the DCT is calculated as- 
\begin{equation}\begin{aligned}F(a,b)=\frac{1}{\sqrt{MN}}\alpha(a)\alpha(b)\sum_{x=0}^{M-1}\sum_{y=0}^{N-1}Q\\ and Q=f(x,y)\times cos(\frac{(2x+1)u\pi}{2M})\times(\frac{(2y+1)v\pi}{2N}) \label{eqn:1}\end{aligned}\end{equation}  where a=0,1,.....M and b=0,1,....N and $\alpha(J)$ is defined by
\[
\alpha(J) =
\begin{cases}
\frac{1}{\sqrt{2}}, & J=0 \\
1, & otherwise
\end{cases}
\]
\subsubsection{Discrete Fourier Transform (DFT)}
\label{subsubsec3}
DFT is a important image processing tool used to separate the spatial domain image into sine and cosine components. In \cite{ASSEFA20101825} and \cite{Jing:2006:FRB:1161716.1161722} we find the application of DFT as feature extraction tool. Like DCT, DFT is also applied on the total skin lesion image and then some of the co-coefficients is used to make the feature vector. The coefficients with global identification capability are taken only. These coefficients encode the low frequency features which are compact and orientation independent. For a $M \times N$ dimensional image DFT is calculated as- 
\begin{equation}F(a,b)=)\sum_{x=0}^{M-1}\sum_{y=0}^{N-1}I(x,y)\times e^{-j\frac{2\pi mx}{M}}e^{-j\frac{2\pi ny}{N}}\label{eqn:2}\end{equation} where $j=\sqrt{-1}$
\subsubsection{Local Binary Pattern (LBP)}
\label{subsubsec4}
\begin{figure}[ht!]
	\includegraphics[width=\linewidth]{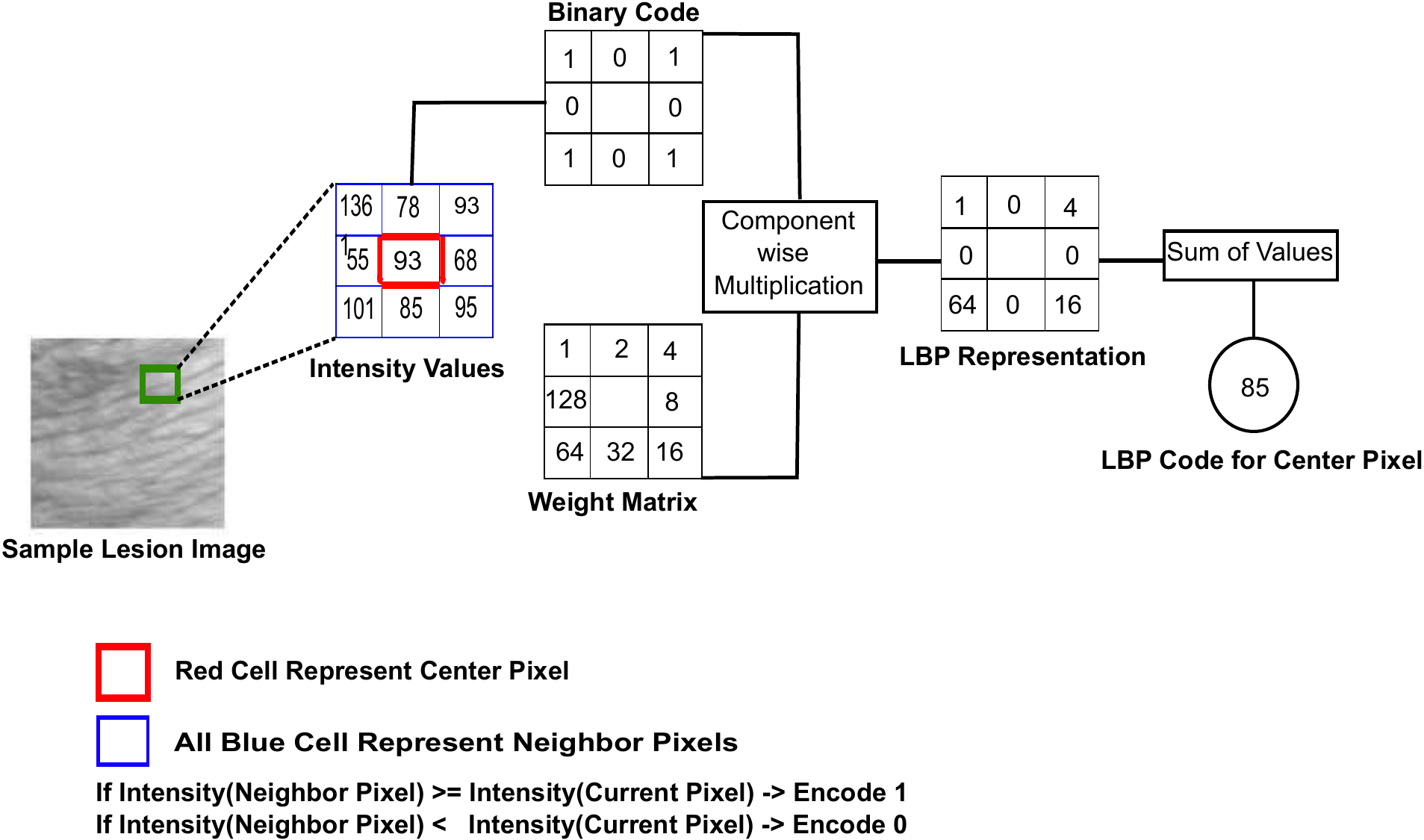}
	\caption{Working principle of LBP using $3\times3$ neighborhood}
	\label{fig:2}
\end{figure}

\begin{figure*}[ht!]
	\centering
	\subfloat[ Leprosy]{{\includegraphics[width=6cm]{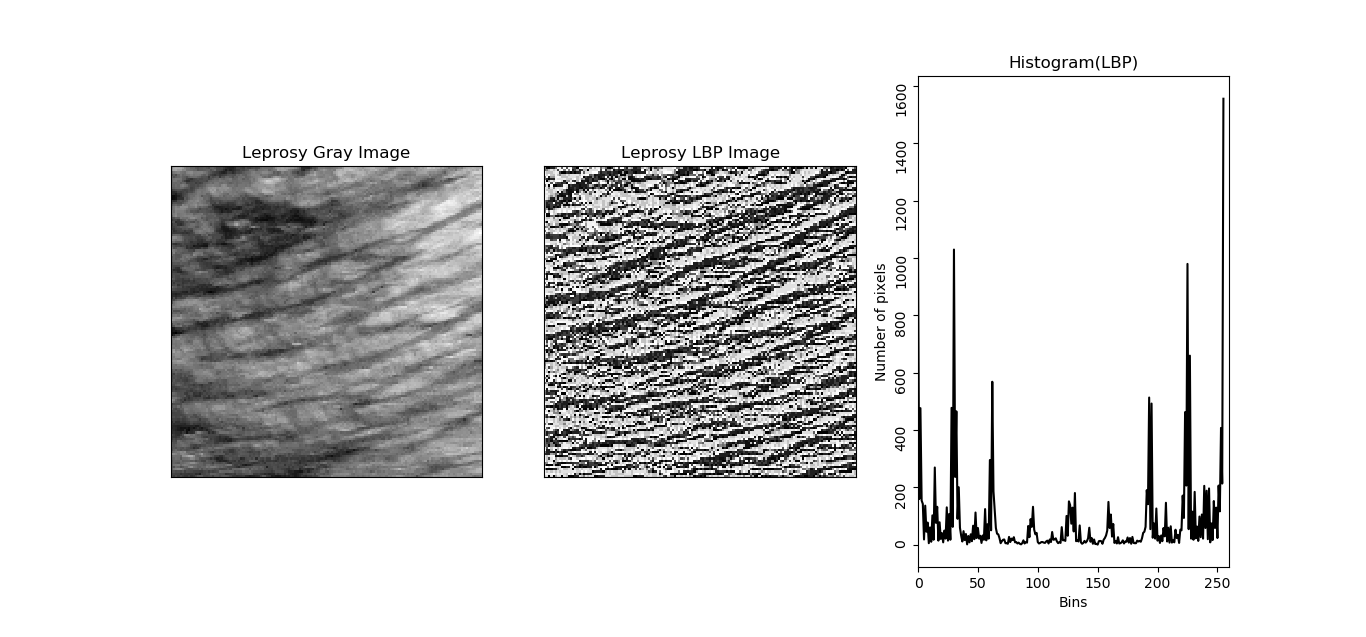} }}%
	\qquad
	\subfloat[Tinea versicolor]{{\includegraphics[width=6cm]{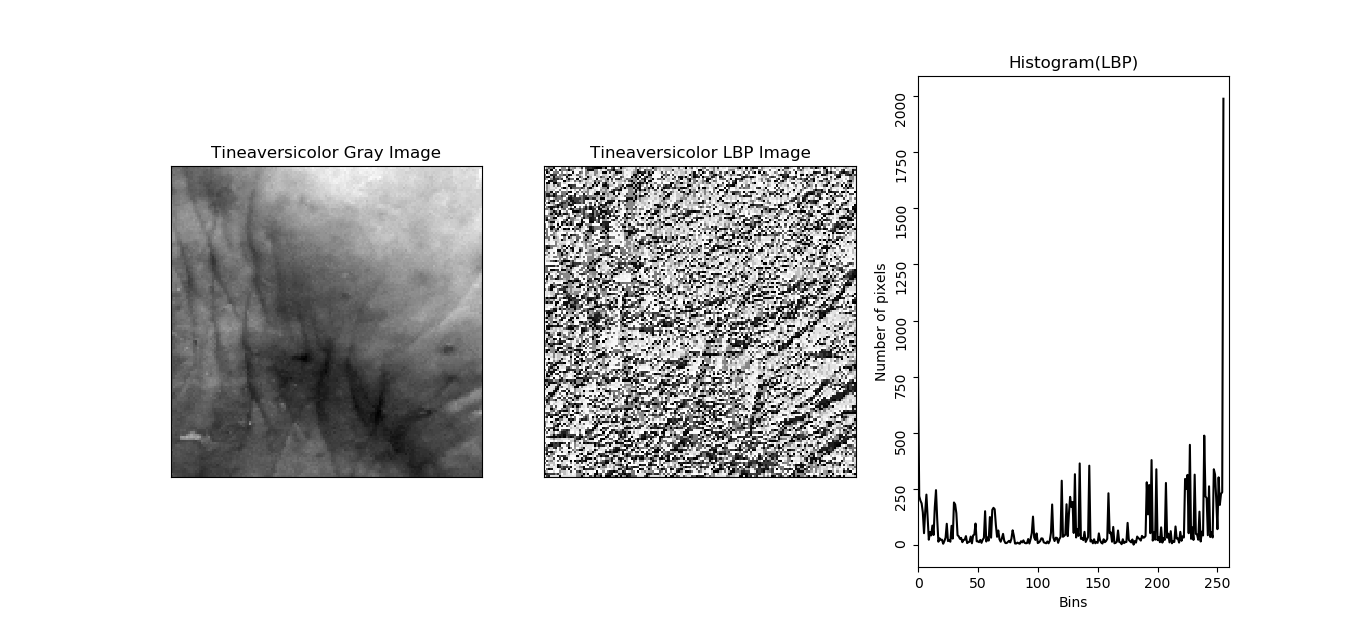} }}%
	\qquad
	\subfloat[ Vitiligo]{{\includegraphics[width=6cm]{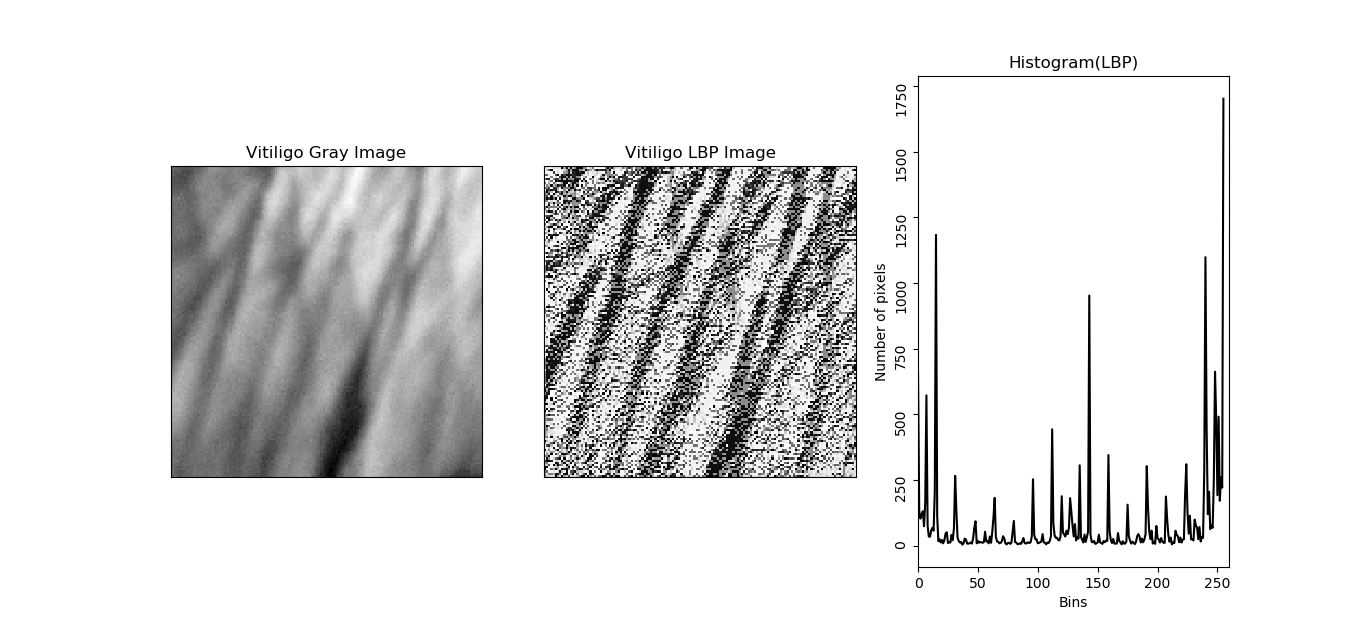} }}%
	\qquad
	\subfloat[Normal]{{\includegraphics[width=6cm]{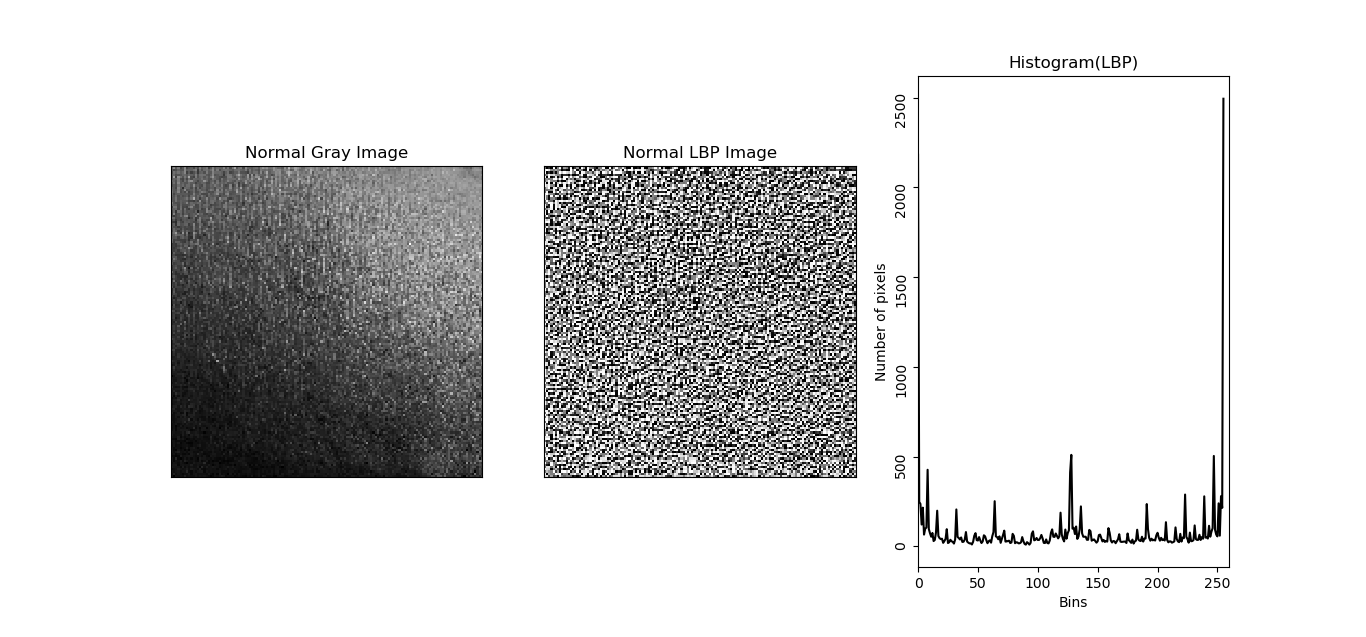} }}%
	\caption{LBP image and histogram of the sample image of every class}%
	\label{fig:3}%
\end{figure*}

Local Binary Pattern is a descriptor, which uses binary derivatives of a pixel to describe the surrounding relationship of that pixel. It generates a bit-code from the binary derivatives of a pixel. Here we represent the bit-code generation principles by LBP using $3 \times 3$ neighborhood pixels. If the neighborhood pixels have greater or equal intensity than the center pixel then 1 will be coded and in the other case 0 will be coded. In such a way a 8-bit binary coded is generated and then depending on the weights a decimal value will be generated. We illustrate this in the figure~\ref{fig:2}.
While calculating LBP code, the edges with less information’s are ignored and thus avoiding false information. Although, we have explained the LBP descriptor with only eight pixels in $3 \times 3$ neighborhood, it can also be computed with different radius and different number of neighboring pixels. By equation~\ref{eqn:3} we can understand this concept. Mathematically the traditional LBP can be expressed by equation~\ref{eqn:4}.
\begin{equation} [x_{p},y_{p}]=[x_{c}+Rcos(\frac{2\pi p}{P}),y_{c}+Rsin(\frac{2\pi p}{P})] \label{eqn:3}\end{equation}
\begin{equation}LBP_{P,R}=\sum_{p=0}^{P-1}sign(g_{p}-g_{c}.2^{p}) \label{eqn:4}\end{equation} where P is the number of perimeter pixels and R is the radius of the neighborhood. Here $g_{p}$  represent the intensities of the surrounding pixels and $g_{c}$ represent the intensities of the center pixel. The sign(x) function is represented by \[ sign(x) =
\begin{cases}
1, & x\geq 0 \\
0, & otherwise
\end{cases}
\]. The LBP operator can be made rotation invariant by rotating the bit-code P-1 times and then choose the minimum decimal equivalent value. We represent the rotation invariant LBP using  $LBP_{P,R}^{ri}$. Mathematically $LBP_{P,R}^{ri}$ can be defined by equation~\ref{eqn:5}.
\begin{equation}LBP_{P,R}^{ri}=min{ROR(LBP_{P,R},i)} \label{eqn:5}\end{equation}
Among all the $LBP_{P,R}^{ri}$ patterns, some of the patterns are carrying the fundamental properties of the image called the uniform patterns. Depending on the number of transitions from 0 to 1 or vice versa the uniform LBP patterns can be detected. If the number of such transition is less than or equal to 2 then the pattern is uniform pattern else non-uniform. The equation~\ref{eqn:6} formally depict the relationship-
\begin{equation}
\resizebox{0.48\textwidth}{!}{
	$LBP_{P,R}^{riu2}=
	\begin{cases}
	\sum_{p=0}^{P-1}sign(g_{p}-g_{c}), & if U(LBP_{P,R})\leq 2\\
	P+1, & otherwise
	\end{cases}$}
\label{eqn:6}\end{equation}
Here $ U(LBP_{P,R})$ can be computed as,
\begin{equation}\resizebox{0.48\textwidth}{!}{$U(LBP_{P,R})= |s(g_{p-1}-g_{c})-s(g_{0}-g_{c}) |+ \sum_{p=1}^{P-1} |s(g_{p}-g_{c})-s(g_{p-1}-g_{c})|$}\label{eqn:7}\end{equation}
LBP has been extensively applied in different pattern recognition and computer vision tasks like \cite{1717463,1530069,1890178,74549,SHAN2009803,5634504,11760023,11760023_29,NANNI20083461,1410446,kundu_local}. We have applied the rotation invariant uniform local binary pattern on the CMATER Skin dataset. From the figure~\ref{fig:3} it is very clear we can distinguish the LBP histogram of different skin lesion images. As there is a certain variance of different LBP histogram, we have applied the LBP for feature extraction in skin diseases identification task.
\subsubsection{Weber Local Descriptor (WLD)}
\label{subsubsec5}
Ernest Weber bases WLD on Weber’s Lawr. Weber’s law establishes a constant relationship between incremental threshold and background intensity. Formally, the Weber’s law can be expressed by $\frac{\Delta I}{I}=K$, where $\Delta I$ is the incremental threshold and I is the initial stimulus intensity. The $\frac{\Delta I}{I}$ is called Weber fraction. WLD is the combination of two components, Differential Excitation and Orientation. The micro variations between the pixels can be encoded by the WLD. 
\begin{itemize}
	\item Differential Excitation:\\
	Using differential excitation we can encode the micro variations between the intensity values of neighboring pixels. Formally, it can be expressed as, 
	\begin{equation}\Delta I=\sum_{i=0}^{p-1}\Delta I(x_i)=\sum_{i=0}^{p-1}I(x_i)-I(x_c), \label{eqn:8}\end{equation} where $i^th$ neighbors of $x_c$ is represented by $x_i(i=0,1,...... p-1)$ and p represents total number of neighbors in a region. $I(x_i)$ presents the intensity of the neighboring pixels, and $I(x_c)$ presents the intensity of current pixel. It can be expressed as,\begin{equation}\xi(x_c)=arctan(\frac{\Delta I}{I})=arctan(\sum_{i=0}^{p-1}(\frac{I(x_i)-I(x_c)}{I(x_i)})).
	\label{eqn:9}
	\end{equation}
	If $\xi(x_c)$ is positive, then center pixel is darker respect to the neighbor pixels and if $\xi(x_c)$ is negative, then current pixel is lighter respect to the neighbor pixels.\
	\item Orientation:\
	It determines the directional property of the pixels. It can be computed as: \begin{equation}\theta(x_c)=arctan(\frac{dI_h}{dI_v}),
	\label{eqn:10}
	\end{equation}
	where $dI_h=I(x_7)-I(x_3)$ and $dI_v=I(x_5)-I(x_1)$  is calculated from the two filters as in the Figure~\ref{fig:2}. The mapping of $f:\theta   \rightarrow \theta ^ \prime$ can be expressed as,
	$\theta ^\prime=arctan2(dI_h,dI_v)+\pi$, and\begin{equation} f(x)=\left\{
	\begin{array}{c l}	
	\theta, & dI_h>0\quad and\quad dI_v>0\\
	\pi - \theta ,& dI_h>0 \quad and \quad dI_v<0\\
	\theta - \pi,& dI_h<0 \quad and \quad dI_v<0 \\
	-\theta,& dI_h<0 \quad and \quad dI_v>0,
	\end{array}\right.
	\label{eqn:11}
	\end{equation}
	where $\theta$ varies from  -$90^\circ$. The quantization is done using,
	\begin{equation}
	\phi_t=f_q(\theta^\prime)=\frac{2t}{T}\pi, where\, t=mod(\left\lfloor \frac{\theta^\prime}{\frac{2\pi}{T}}+\frac{1}{2} \right\rfloor,T).
	\label{eqn:12}
	\end{equation}
	The working principle of WLD in a $3\times3$ neighborhood is easily depicted in figure~\ref{fig:4}.  Again, we can make the WLD rotation invariant by calculating the orientation component in all the eight directions and take the minimum value of it. The differential excitation component is inherently rotation invariant because we compute the difference of all neighboring pixel with the center pixel. Formally the rotation invariant WLD (WLDRI) can be expressed by 
	\begin{equation}\label{eqn:13}\resizebox{.9\hsize}{!}{$\theta_i=arctan(\frac{I(x_((\frac{p}{2})-i)mod\, p)-I(x_i)}{I(x_((\frac{3p}{4})-i)mod\, p)-I(x_((\frac{p}{4})-i)mod\, p)}), and \quad \theta(x_c)=min_{i=0}^{p-1}(\theta_i)$}\end{equation}
	\begin{figure}[ht!]
		\includegraphics[width=\linewidth]{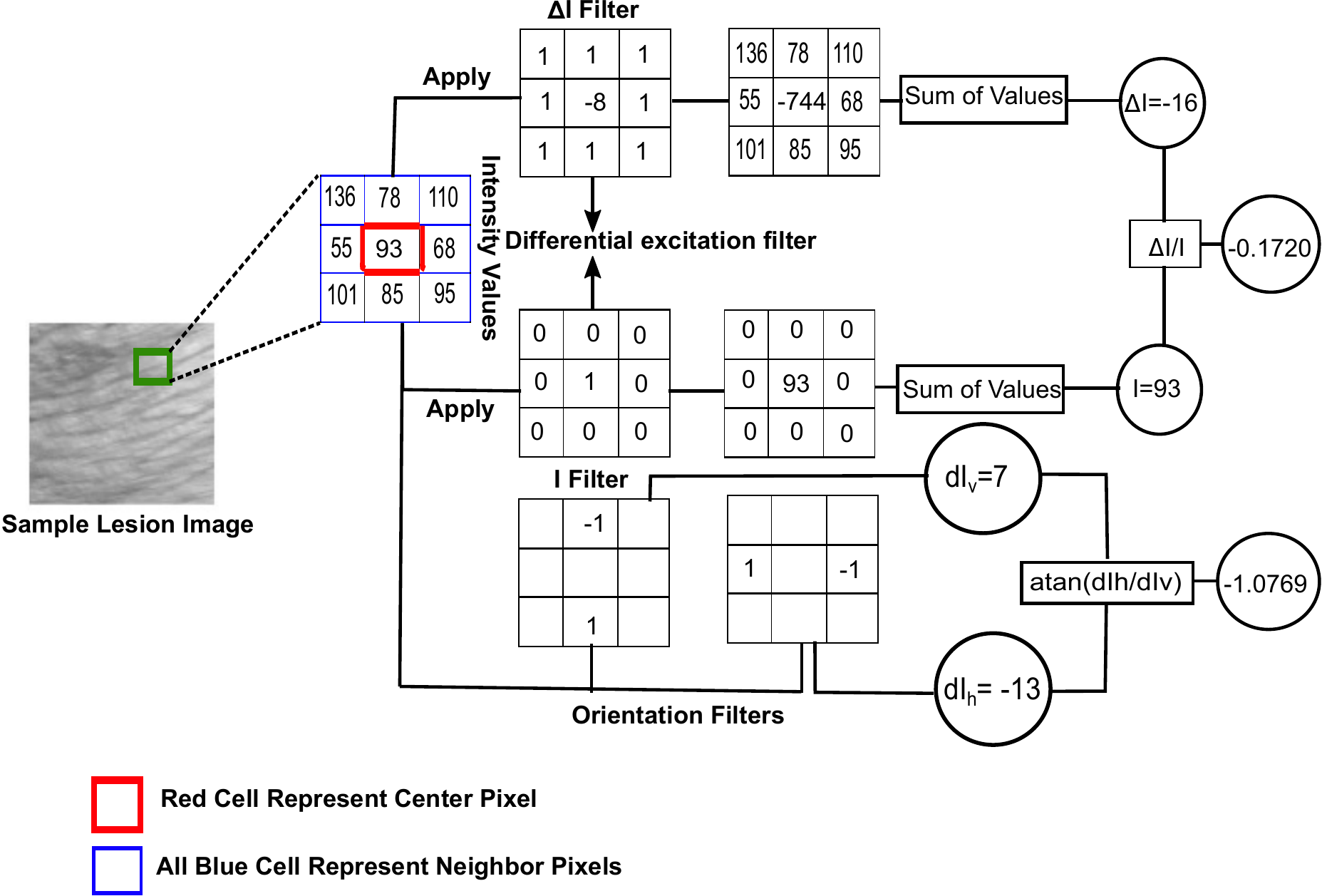}
		\caption{Illustration how WLD works using $3\times3$ neighborhood}
		\label{fig:4}
	\end{figure}
	\item WLD Histogram:\\
	We can construct the WLD Histogram separately for the above two components, namely, differential excitation and orientation. Both differential excitation and orientation can take the values in the range of $\frac{-\pi}{2}$ to $\frac{\pi}{2}$.  The values of differential excitation are quantized into T equal width bins and values of orientation are quantized into M equal width bins. Using differential excitation and orientation we form the 2D histogram $h_{T,M}$. When the pixels have the same differential excitation and orientation, then the pixels will put into same bin. Transforming the 2D histogram into a 1D histogram creates the feature vector. In our work we used T=6 and M=8. So, the size of the feature vector in our case is 48. We can visualize the effect of applying WLD on the skin diseases images and the histogram of the differential excitation in the range from 0-255 using the figure~\ref{fig:5}. WLD clearly highlights the strong edges present in the image which will be useful in the recognition task. 
	\begin{figure}[ht!]
		\centering
		\subfloat[ Leprosy]{{\includegraphics[width=3.5cm]{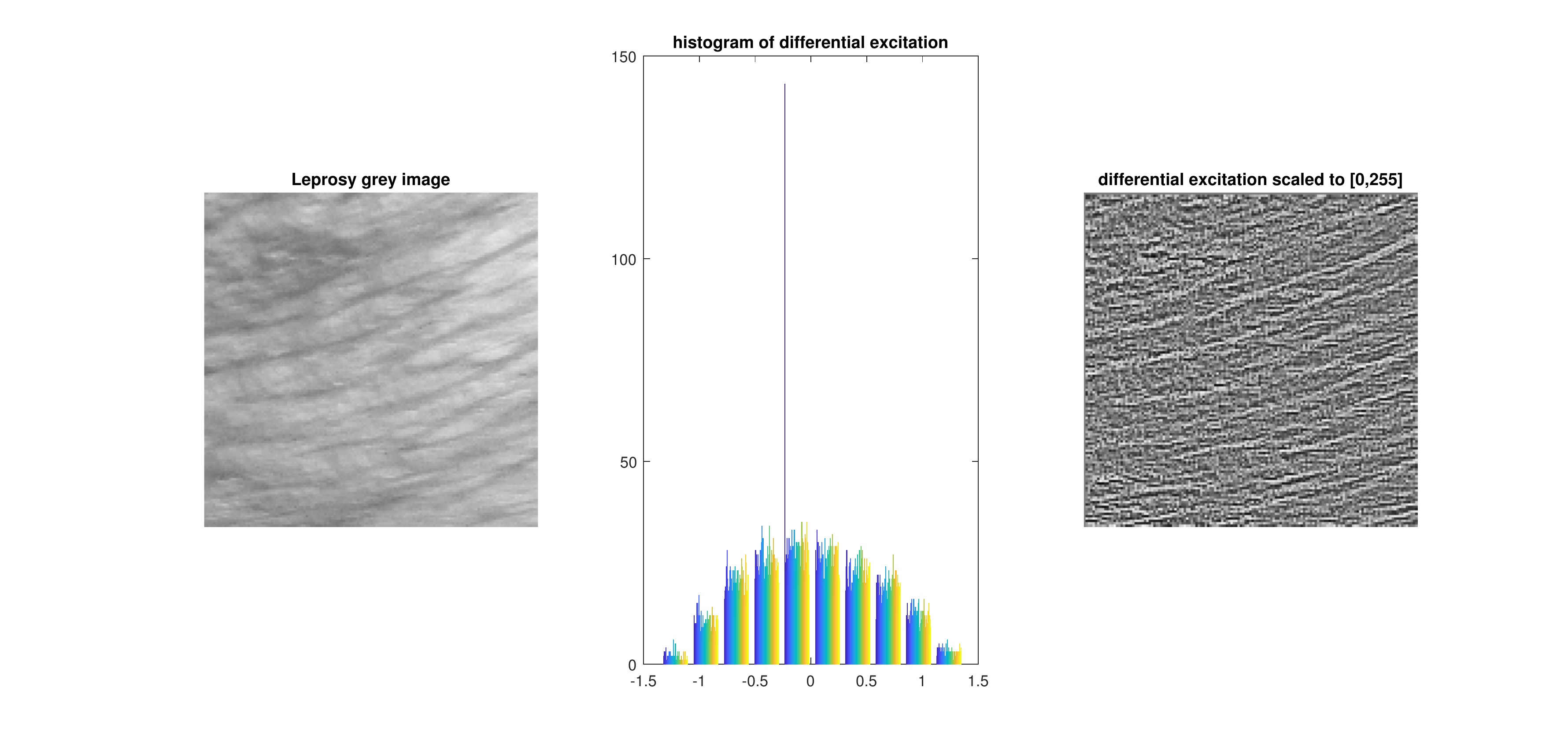} }}%
		\qquad
		\subfloat[Tineaversicolor]{{\includegraphics[width=3.5cm]{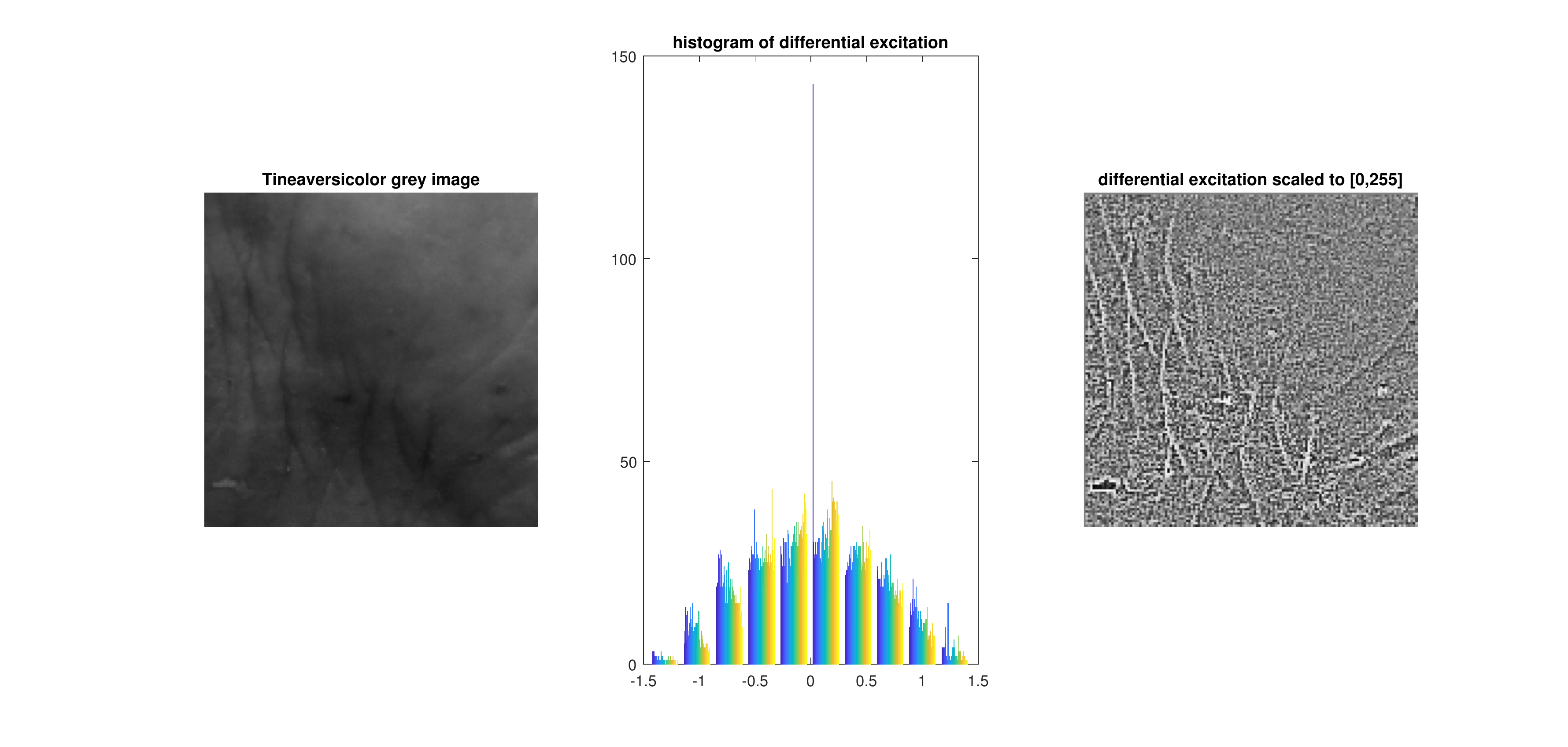} }}%
		\qquad
		\subfloat[ Vitiligo]{{\includegraphics[width=3.5cm]{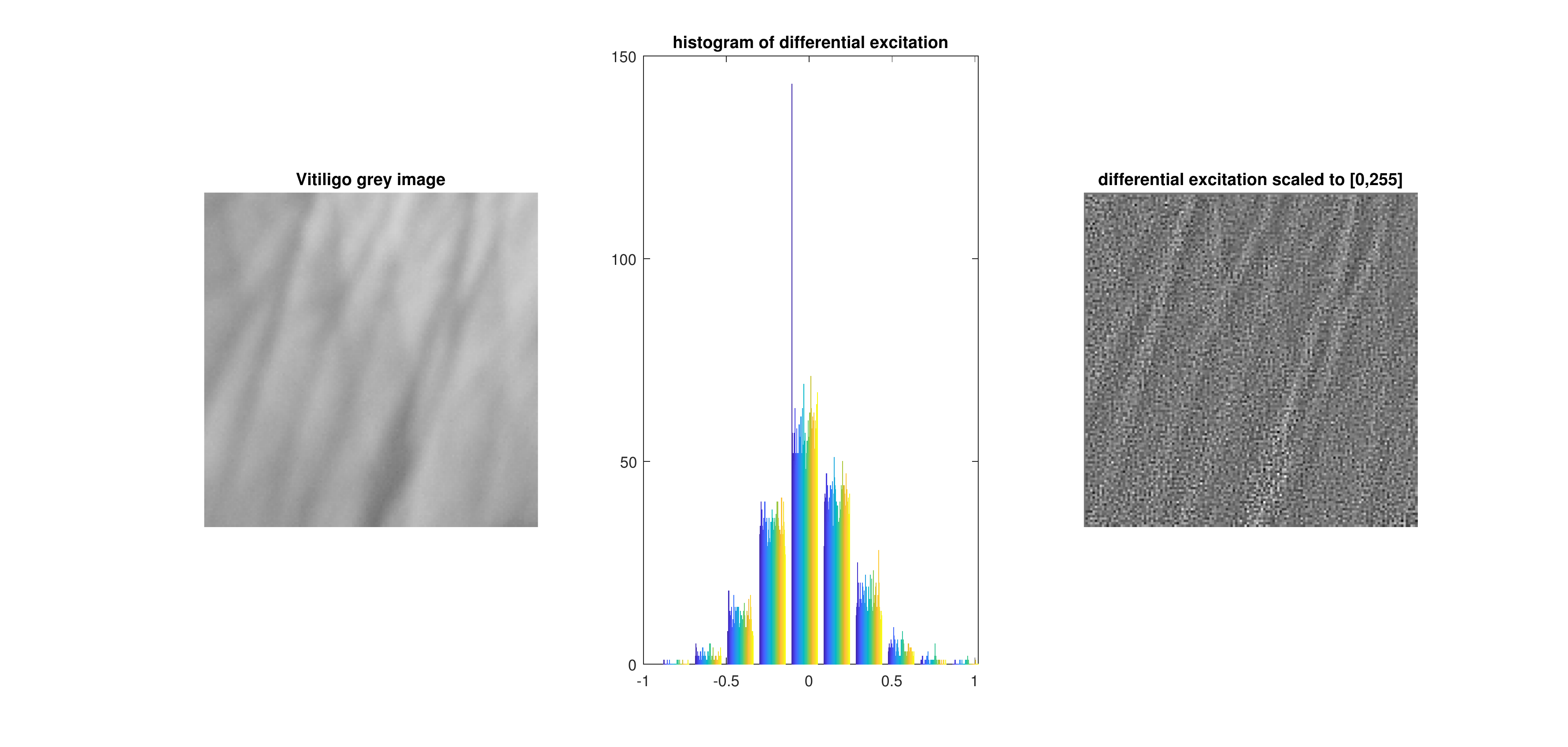} }}%
		\qquad
		\subfloat[Normal]{{\includegraphics[width=3.5cm]{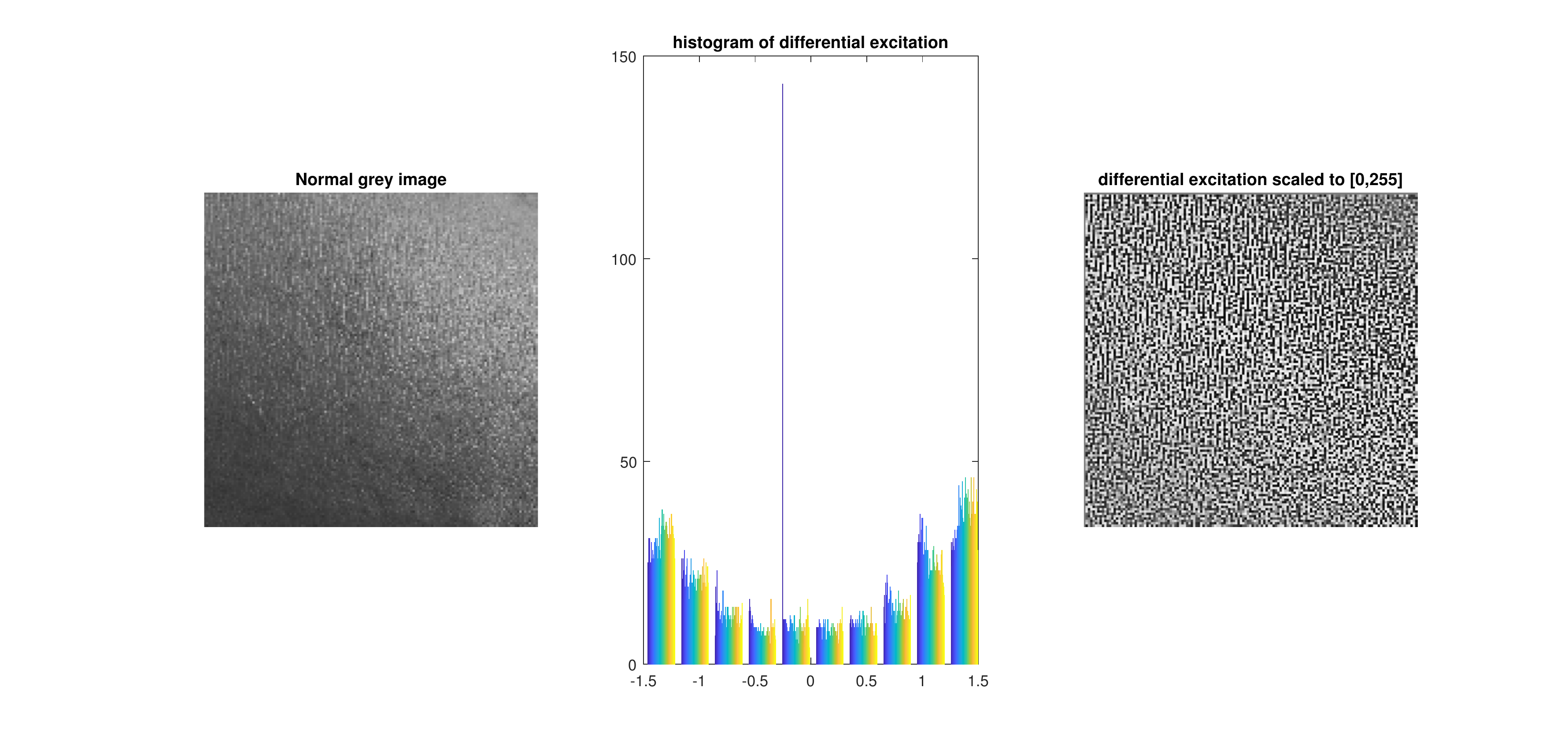} }}%
		\caption{ Histogram and scaled map of differential excitation}
		\label{fig:5}%
	\end{figure}
\end{itemize}
\subsection{Experimental Setup and Results}
\label{subsec4}
As stated earlier, we are using CMATER Skin dataset for the experiment which consists images of three skin diseases Leprosy, Tinea versicolor and Vitiligo. Also, the normal skin images have been considered to distinguish them from the other three skin disease images. In CMATER Skin dataset a total of 876 images of normalized size of $144 \times 144$ are available. We have divided the dataset into 4:1 ratio for preparing the training and testing datasets. The overall methodology is described in the figure~\ref{fig:1}. At first, the region of interest is manually cropped from the camera captured skin lesion image. The only pre-processing employed is to convert the color lesion image to a gray scale image. We have applied both rotation invariant WLD and rotation invariant LBP in this work to get the best micro textural pattern and image gradients from the images in a best competitive environment. Several experiments are done for the skin diseases recognition task to achieve the best set of results.  
\begin{itemize}
	\item
	LBP and WLD without rotation invariance is applied on the CMATER Skin dataset. Different P and R values are used $([P=8,R=1],[P=16,R=2],[P=24,R=3])$ for feature extraction using both LBP and WLD. Employment of multi scale approach have been applied to achieve the scale invariance and to get more discriminate features. The same experiment we have done with the rotation invariant WLD and LBP. The rotation invariant features work better when we have to test a unknown rotated skin lesion image. The table~\ref{tab1} gives the feature size.\\
	\begin{table}[ht!]
		\centering
		\renewcommand{\arraystretch}{1.2}
		\caption{Feature vector size using for different (P,R) values}
		\label{tab1}       
		\begin{tabular}{llll}
			\hline
			Feature Descriptor &(8,1)&(16,2)&(24,3) \\ \hline
			LBP &59&243&555\\ 
			WLD&48&48&48\\ \hline
			riLBP &10&18&26\\ 
			WLDRI&48&48&48\\ \hline
		\end{tabular}
	\end{table}
	\item
	Computed features for all (P,R) values are combined together to employ the micro patterns from all the scales. The performance of combined features is better than the single feature set. The feature vector size is depicted in table~\ref{tab3}.
	\begin{table}[ht!]
		\centering
		\renewcommand{\arraystretch}{1.2}
		\caption{Feature vector size of combined feature set}
		\label{tab2}       
		\begin{tabular}{ll}
			\hline
			Feature Descriptor &(8,1)+(16,2)+(24,3) \\ \hline
			LBP &857\\  
			WLD&144\\ 
			riLBP & 54\\ 
			WLDRI & 144\\ \hline
		\end{tabular}
	\end{table}
	\item
	The image is divided into four regions based on the center of gravity(cog) and the multi scale features are extracted from the four regions for different (P,R) values. Here we get the more discriminate feature set for distinguishing the skin diseases. This approach is applied for both LBP and WLD. The feature vector size is depicted in table~\ref{tab3}.
	\begin{table}[ht!]
		\centering
		\renewcommand{\arraystretch}{1.2}
		\caption{Feature vector size for center of gravity based approach}
		\label{tab3}       
		\begin{tabular}{llll}
			\hline
			Feature Descriptor &(8,1)&(16,2)&(24,3) \\ \hline
			cogLBP &236&972&2220\\  
			cogWLD&192&192&192\\ 
			cogriLBP &40&72&104\\ 
			cogWLDRI &192&192&192\\ \hline
		\end{tabular}
	\end{table}
	\item
	The center of gravity based features using WLD and LBP are combined and tested on the skin diseases recognition task. The feature vector size is depicted in table~\ref{tab4}.
	\begin{table}[ht!]
		\centering
		\renewcommand{\arraystretch}{1.2}
		\caption{Feature vector size for combined center of gravity based features}
		\label{tab4}       
		\begin{tabular}{ll}
			\hline
			Feature Descriptor &(8,1)+(16,2)+(24,3) \\ \hline
			cogLBP & 3428\\  
			cogWLD&576\\ 
			cogriLBP &216\\ 
			cogWLDRI &576\\ \hline
		\end{tabular}
	\end{table}
	\item
	We have combined the features of LBP and WLD to make a combined feature which will extract the micro patters as well as preserve the strong edge gradients of the image. The combination of LBP and WLD makes a very strong feature vector by which the system learns optimally and produce promising recognition accuracy. The feature vector size is depicted in table~\ref{tab5}.
	\begin{table}[ht!]
		\centering
		\renewcommand{\arraystretch}{1.2}
		\caption{Feature vector size for combination of LBP and WLD}
		\label{tab5}       
		\begin{tabular}{llll}
			\hline
			Feature Descriptor &(8,1)&(16,2)& (24,3) \\ \hline
			WeberLBP &117&291&603 \\  
			riWeberLBP&58&66&74\\ \hline
		\end{tabular}
	\end{table}
	\item
	Computed features from the combination of LBP and WLD for all (P,R) values are combined together and used for the experiment. The feature vector size is depicted in table ~\ref{tab6}
	\begin{table}[ht!]
		\centering
		\renewcommand{\arraystretch}{1.2}
		\caption{Feature vector size for WeberLBP based features}
		\label{tab6}       
		\begin{tabular}{ll}
			\hline
			Feature Descriptor &(8,1)+(16,2)+(24,3) \\ \hline
			cogWeberLBP & 1011\\  
			cogriWeberLBP&198\\ \hline
		\end{tabular}
	\end{table}
	\item
	Based on center of gravity the WeberLBP and riWeberLBP features are computed. The feature vector size is depicted in table~\ref{tab7}
	\begin{table}[ht!]
		\centering
		\renewcommand{\arraystretch}{1.2}
		\caption{Feature vector size for  center of gravity based WeberLBP features}
		\label{tab7}       
		\begin{tabular}{llll}
			\hline
			Feature Descriptor &(8,1)&(16,2)&(24,3) \\ \hline
			WeberLBP & 468& 1164 & 2412\\  
			riWeberLBP&232 & 264 & 296\\ \hline
		\end{tabular}
	\end{table}
	\item
	The center of gravity based WeberLBP and riWeberLBP features are combined together to form enriched micro pattern and edge gradient based strong feature set. The feature vector size is depicted in table~\ref{tab8}
	\begin{table}[ht!]
		\centering
		\renewcommand{\arraystretch}{1.2}
		\caption{Feature vector size for combined center of gravity based  WeberLBP features}
		\label{tab8}       
		\begin{tabular}{ll}
			\hline
			Feature Descriptor &(8,1)+(16,2)+(24,3) \\ \hline
			WeberLBP & 4044\\  
			riWeberLBP&792\\ \hline
		\end{tabular}
	\end{table}
\end{itemize}
While developing the system for skin disease identification system, our main objective is to identify the feature set which provides the best recognition result. Because these types of real-life applications need better accuracy to be useful in real cases. The working methodology and different experimental protocols are described in the following sections. The features acquired under different experiment protocols are used to train a Multi-SVM classifier. The grid search method has been applied to find the best regularization parameter(C) and kernel coefficient of rbf kernel(gamma). The unknown images are classified by the trained model of SVM. Here we are using different variations of LBP for all the experiments using different property and combinations with WLD. The results using different experiment protocols are described in the following.
\begin{itemize}
	\item{\textbf {Experiment Protocol 1:}}
	In this experiment protocol we use traditional LBP in the neighborhood of (8,1), (16,2) and (24,3) to acquire the features from different scales. We have achieved 74.1379 \%, 73.5632\% and 73.5632\% accuracy using (8,1), (16,2) and (24,3) neighborhoods respectively. 
	\item{\textbf {Experiment Protocol 2:}}
	The Rotation invariant LBP(riLBP) is used here. We have used three different neighborhood (8,1), (16,2) and (24,3) to acquire the features from different scales. Using riLBP we have achieved 87.931\%, 89.0805\% and 87.931\% accuracy using (8,1), (16,2) and (24,3) neighborhoods respectively. We can see the significant improvement because of involvement of rotation invariant features in LBP. In rotation invariant LBP, the feature vector size reduces significantly as we use uniform LBP patterns only. Due to uniform rotational invariant features the recognition accuracy increases significantly.  
	\item{\textbf {Experiment Protocol 3:}}
	The WLD and LBP features are combined together to acquire the strong feature set called WeberLBP, used to classify the images associated with three skin diseases. Like the previous two experiment three different neighborhood is used. Using WeberLBP we have achieved 82.1839\%, 81.6092\% and  86.2069\% accuracy using (8,1), (16,2) and (24,3) neighborhoods respectively. We can see that using the enriched combination of WLD and LBP 8.046\%, 8.046\% and 12.6437\% improvements in the recognition accuracies over those obtained with traditional LBP have been achieved.
	\item{\textbf {Experiment Protocol 4:}}
	This experiment uses the combination of riLBP and WLDRI to make a strong rotation invariant feature set(riWeberLBP). We get 87.931\%, 89.08085\% and 89.0805\% accuracy using (8,1), (16,2) and (24,3) neighborhood. Clearly 1.1495\% improved accuracy can be seen in (24,3) neighborhood over riLBP using riWeberLBP.
	\item{\textbf {Experiment Protocol 5:}}
	The image is divided into four parts using center of gravity and from each part the traditional LBP is applied. Using this approach we can acquire more detailed information of the sub regions of the image. We get 79.88\%, 75.2874\% and 75.2874\% accuracy using (8,1), (16,2) and (24,3) neighborhoods respectively by applying cogLBP. This approach gives 5.7421\%, 1.7242\% and 1.7242\% improved accuracy over traditional LBP.
	\item{\textbf {Experiment Protocol 6:}}
	Here the rotation invariant LBP is applied on the sub regions of the image based on center of gravity. Using this approach we have achieved 86.7816\%, 90.8046\% and 87.3563\% accuracy using (8,1), (16,2) and (24,3) neighborhoods respectively. Clearly we have achieved 1.724\% improved accuracy over traditional LBP on (16,2) neighborhood using cogriLBP.
	\item{\textbf {Experiment Protocol 7:}}
	In this experiment, the WLD and LBP is applied on the sub regions of the image. Using the center of gravity, the image is divided into four sub regions. The cog based WLD and LBP features are combined together to get feature set cogWeberLBP. In this approach we have achieved 81.6092\%, 81.0345\% and 86.78\% accuracy using (8,1), (16,2) and (24,3) neighborhoods respectively.
	\item{\textbf {Experiment Protocol 8:}}
	Here rotation invariant WLD and LBP is applied on the sub regions of the image and the features are combined to make the strong feature set cogriWeberLBP. Using cogriWeberLBP we have achieved 88.5057\%, 91.3793\% and 88.5057\% accuracy using (8,1), (16,2) and (24,3) neighborhoods respectively. Clearly cogriWeberLBP achieve the highest accuracy using on (16,2) neighborhood.
	\item{\textbf {Experiment Protocol 9:}}
	The features associated with three different scales are combined together for the LBP, riLBP, WeberLBP, riWeberLBP, cogLBP, cogriLBP, cogWeberLBP, cogriWeberLBP. The accuracies using combinedLBP and combinedriLBP are 81.339\% and 81.7664\% respectively. When we combine WeberLBP and riWeberLBP features for all the mentioned scales, 83.0484\% and 85.4701\% ac curacies are obtained.
	\item{\textbf {Experiment Protocol 10:}}
	We implement some of the popular deep learning networks like GoogLeNet, MobileNet, ResNet\textunderscore152, DenseNet\textunderscore121, ResNet\textunderscore101 to compare the results with the proposed method. Here we used a random flip to increase the size of the dataset. ResNet\textunderscore101 gives better result (89\%) than the other used deep learning networks. The results are tabulated in table~\ref{tab10}.
\end{itemize}

\begin{figure}[ht!]
	\centering
	\subfloat[Leprosy Detected as Tinea Versicolor]{{\includegraphics[width=2cm]{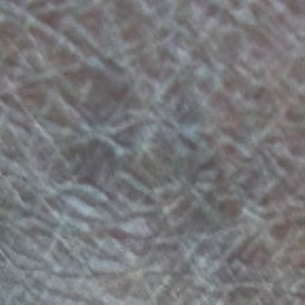} }}%
	\qquad
	\subfloat[Leprosy Detected as Vitiligo]{{\includegraphics[width=2cm]{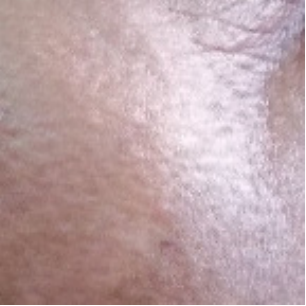} }}%
	\qquad
	\subfloat[Normal Detected as Leprosy]{{\includegraphics[width=2cm]{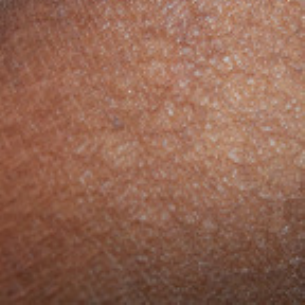} }}%
	\qquad
	\subfloat[Tinea Versicolor detected as Leprosy]{{\includegraphics[width=2cm]{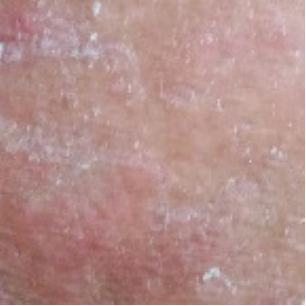} }}%
	\qquad
	\subfloat[Vitiligo detected as Leprosy]{{\includegraphics[width=2cm]{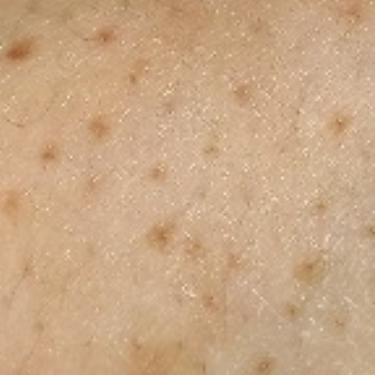} }}%
	\caption{Some examples of incorrectly detected skin lesion image}
	\label{fig:7}%
\end{figure}

We get the best accuracy of 91.3793\% using center of gravity based WeberLBP (cogWeberLBP) on (16,2) neighborhood. The results of different experiment protocols with the best SVM parameters are tabulated in table~\ref{tab9}. We have also compared the results of proposed approach on CMATER skin dataset with the GLCM, DFT and DCT based methods\cite{6686372}. The DCT coefficient based method achieved 73.56\% accuracy using Nu value of 0.080 and gamma value of 0.059. The result of DFT coefficient based method is not promissible compared to DCT coefficient based method. The DFT coefficient based method achieved 70.69\% accuracy using Nu value of 0.045 and gamma value of 0.08. The GLCM based method achieved 88.51\% accuracy on CMATER skin dataset using Nu value of 0.120 and gamma value of 0.070. Whereas center of gravity based rotation invariant  WeberLBP(cogriWeberLBP) achieved 91.38\% accuracy which is significantly better than the GLCM, DFT and DCT based method.
\begin{figure}[ht!]
	\centering
	\subfloat[Results of different Proposed Methods]{{\includegraphics[width=3.5cm]{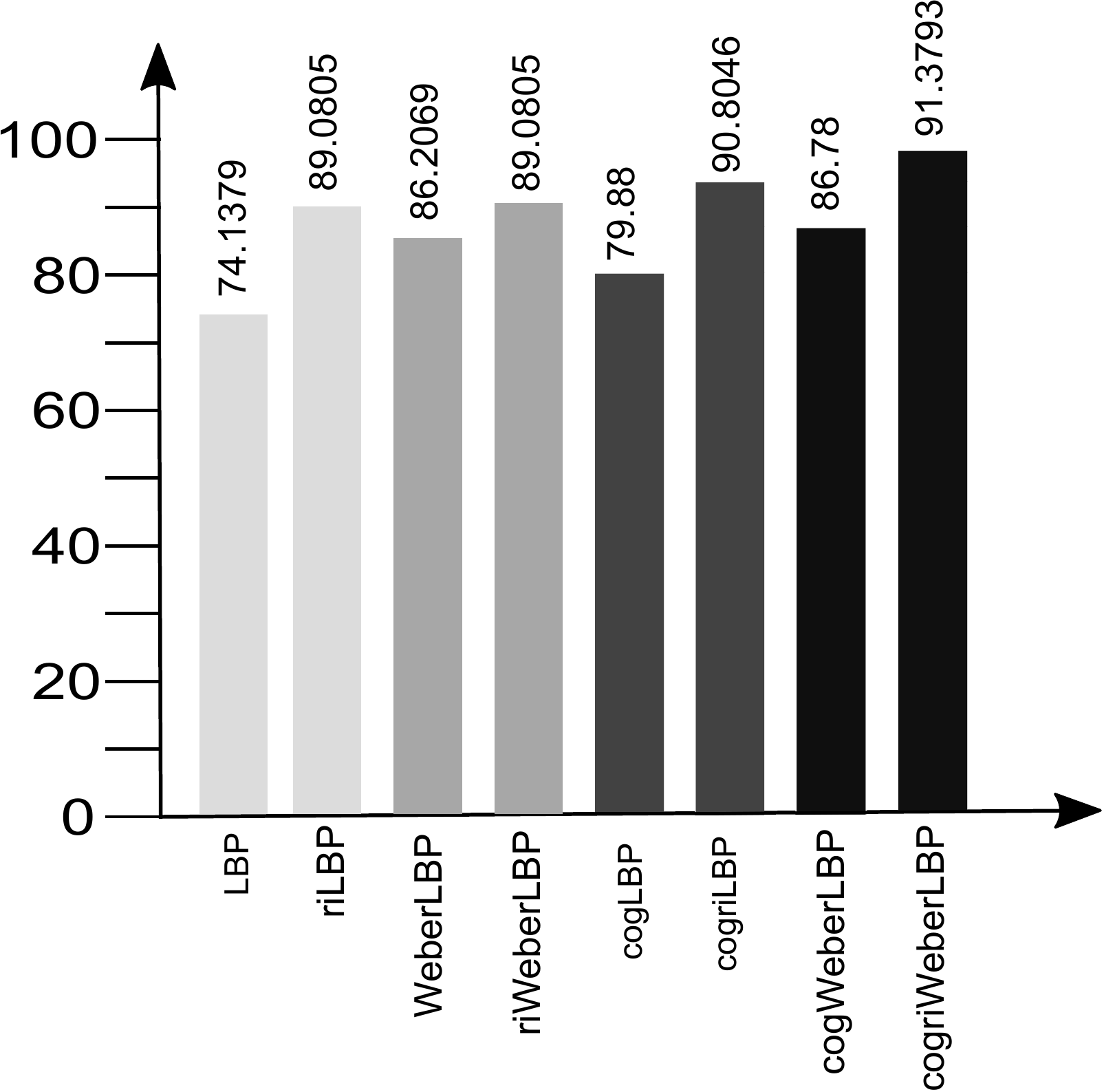} }}%
	\qquad
	\subfloat[Comparison of Proposed Method with DFT, DCT and GLCM]{{\includegraphics[width=3.5cm]{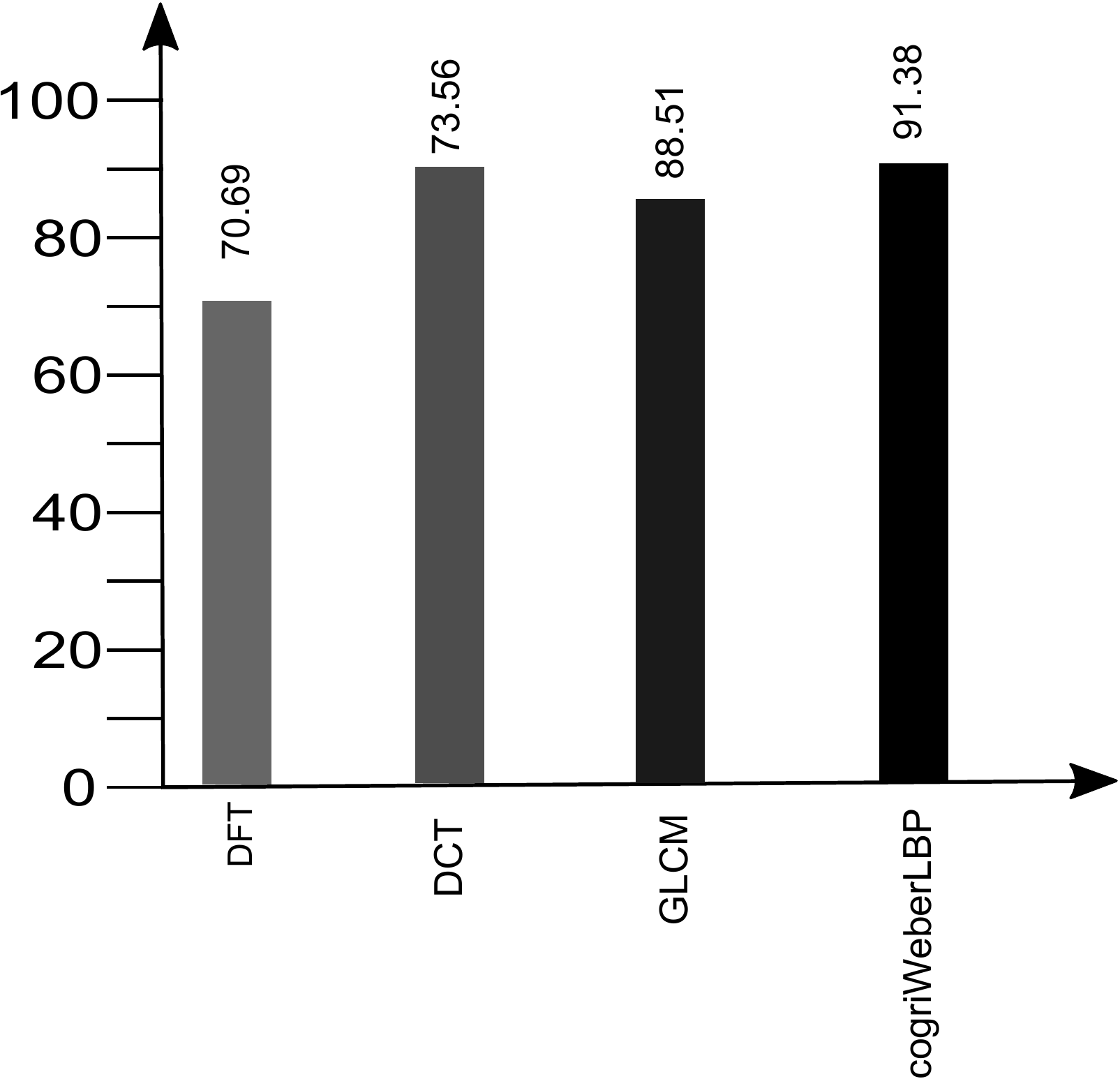} }}%
	\qquad
	\subfloat[Comparison of Proposed Method
	with some popular deep learning networks]{{\includegraphics[width=4cm]{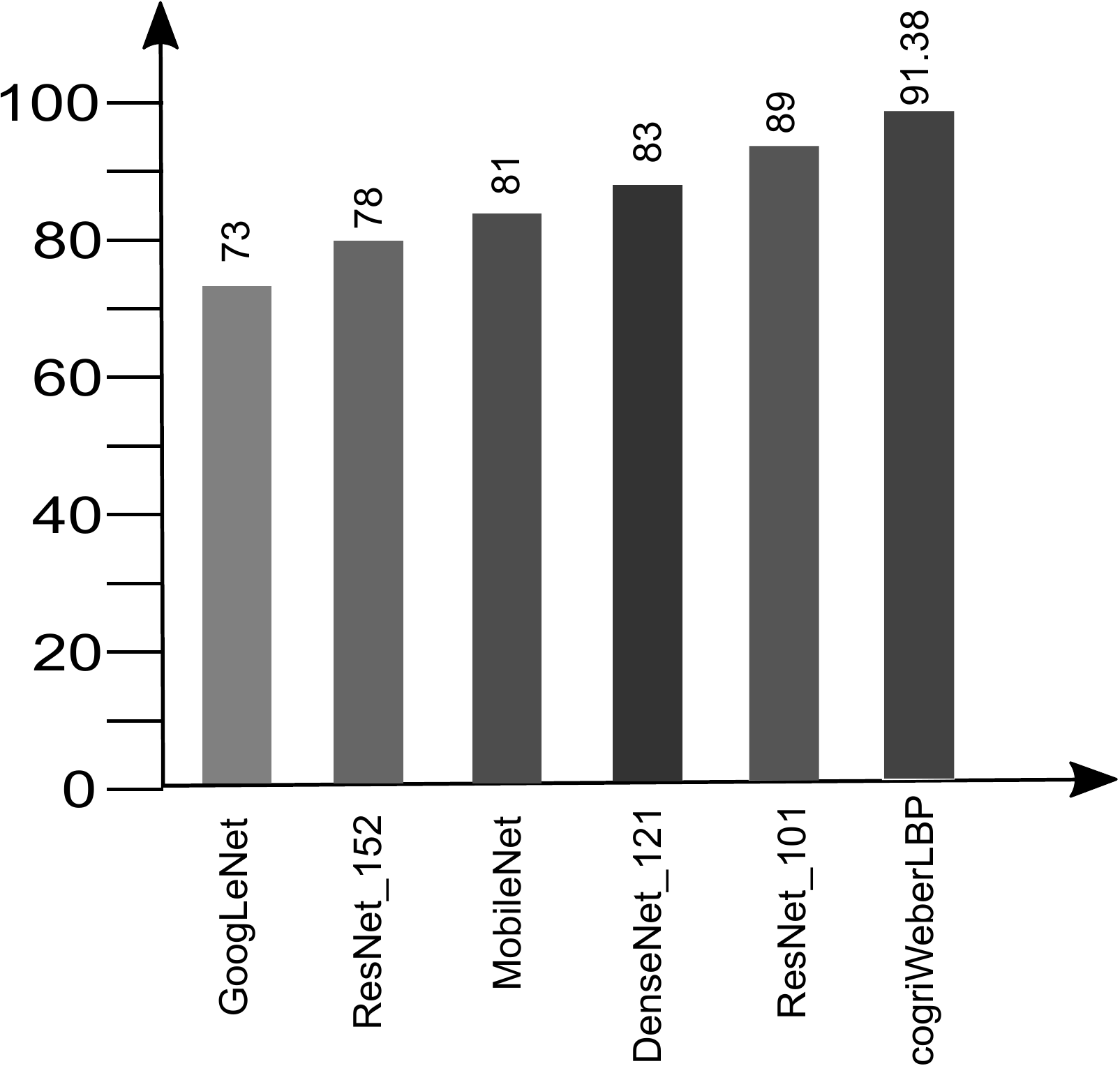} }}%
	\caption{Results of proposed methods and comparison with DCT, DFT, GLCM and some popular deep learning networks}%
	\label{fig:6}%
\end{figure}
The results are compared with some popular deep learning networks such as GoogLeNet, ResNet\textunderscore152, DenseNet\textunderscore121, MobileNet, and ResNet\textunderscore101. A random flip operation is applied on the CMATER skin dataset for data augmentation. We get 89\% accuracy using  ResNet\textunderscore101, whereas using GoogleNet, ResNet\textunderscore152, MobileNet and DenseNet\textunderscore121 we have achieved 73\%, 78\%, 81\%, and 83\% correspondingly. In spite of small intra class variation , we have achieved better accuracy compared to the previous approaches. But for some of the input images the skin diseases identification system failed to give the correct result. In figure~\ref{fig:7} some examples of incorrectly detected skin lesion images are depicted. The images of leprosy affected area with big white patches, are incorrectly detected as vitiligo. The normal skin images with red spots are wrongly detected as leprosy because it is the fundamental symptom of leprosy. The same happens with some of the Tinea versicolor affected skin images and only one of vitiligo affected skin image. 
\begin{table*}[t]
	\centering
	\caption{Results of different experiment protocols with best SVM parameters}
	\label{tab9}       
	\begin{center}
		\resizebox{\textwidth}{!}{
			\begin{tabular}{l|l|l|l|l}
				\hline
				\textbf {Methods}              &                & \textbf{(8,1)}                  & \textbf{(16,2)}                        & \textbf{(24,3)}                      \\ \hline
				\multirow{2}{*}{\textbf{LBP}} & \textbf{Accuracy}       & 74.1379                & 73.5632                       & 73.5632                     \\ \cline{2-5}
				& \textbf{SVM Parameters} & C=512 & C=8192& C=512\\
				&&Gamma=0.007813&Gamma=0.000122  &Gamma=0.000122\\\hline
				\multirow{2}{*}{\textbf{riLBP}} & \textbf{Accuracy}       & 87.931                & 89.0805                       & 87.931                     \\ \cline{2-5}
				& \textbf{SVM Parameters} & C=512 & C=8192& C=32768\\
				&&Gamma=0.007813&Gamma=0.0078125  &Gamma=0.000488\\\hline
				\multirow{2}{*}{\textbf{WeberLBP}} & \textbf{Accuracy}       & 82.1839                & 81.6092                       & 86.2069                     \\ \cline{2-5}
				& \textbf{SVM Parameters} & C=8 & C=2& C=8\\
				&&Gamma=0.5&Gamma=0.125  &Gamma=0.03125\\\hline
				\multirow{2}{*}{\textbf{riWeberLBP}} & \textbf{Accuracy}       & 87.931                & 89.0805                       & 89.0805                    \\ \cline{2-5}
				& \textbf{SVM Parameters} & C=128 & C=8& C=8\\
				&&Gamma=0.5&Gamma=0.5  &Gamma=0.5\\\hline
				\multirow{2}{*}{\textbf{cogLBP}} & \textbf{Accuracy}       & 79.88                & 75.2874                       & 75.2874                    \\ \cline{2-5}
				& \textbf{SVM Parameters} & C=8 & C=8& C=32\\
				&&Gamma=0.125&Gamma=0.007813 &Gamma=0.000488\\\hline                
				\multirow{2}{*}{\textbf{cogriLBP}} & \textbf{Accuracy}      & 86.7816                & 90.8046                       & 87.3563                    \\ \cline{2-5}
				& \textbf{SVM Parameters} & C=2 & C=8& C=8\\
				&&Gamma=2&Gamma=0.5  &Gamma=0.5\\\hline
				\multirow{2}{*}{\textbf{cogWeberLBP}} & \textbf{Accuracy}       & 81.6092                & 81.0345                       & 86.78                    \\ \cline{2-5}
				& \textbf{SVM Parameters} & C=8 & C=8& C=2048\\
				&&Gamma=0.03125&Gamma=0.007813  &Gamma=3.05176e-05\\\hline
				\multirow{2}{*}{\textbf{cogriWeberLBP}} & \textbf{Accuracy}       & 88.5057                & 91.3793                       & 88.5057                    \\ \cline{2-5}
				& \textbf{SVM Parameters} & C=8 & C=8& C=8\\
				&&Gamma=0.125&Gamma=0.125  &Gamma=0.125\\\hline                    
		\end{tabular}}
	\end{center}
\end{table*}
\begin{table}[ht!]
	\centering
	\renewcommand{\arraystretch}{1.2}
	\caption{Comparison of some popular deep learning networks with cogriWeberLBP on the skin diseases dataset}
	\label{tab10}       
	\begin{tabular}{ll}
		\hline
		Methods & Recognition Accuracy(\%) \\ \hline
		GoogLeNet &73\\  
		ResNet\textunderscore152 &78\\ 
		MobileNet & 81\\ 
		DenseNet\textunderscore121 & 83\\
		ResNet\textunderscore101 & 89\\
		cogriWeberLBP & 91.3793\\ \hline
	\end{tabular}
\end{table}
\section{Conclusion}
\label{section4}
In the present work, we have proposed a enrich ensemble of LBP and WLD and applied to identify three similar looking skin diseases- Leprosy, Vitiligo and Tinea versicolor. The cogriWeberLBP gained the best accuracy of 91.38\% on CMATER skin dataset. The proposed method has outperformed the other feature descriptor based approaches applied on the same dataset. The skin disease recognition accuracy using DFT, DCT, GLCM and WLD is 70.60\%, 73.56\%, 88.51\%, 89.08\% respectively. cogriWeberLBP also outperformed some of the most popular deep neural networks such as MobileNet, ResNet\textunderscore152, GoogLeNet, DenseNet\textunderscore121, and  ResNet\textunderscore101. The proposed method clearly achieves significant better result than the other texture based approaches. The proposed system makes the system very robust to recognize the diseases without the involvement of the specialized doctors and thus very easy to get the faster treatment. This system is specially helpful in the rural areas where good dermatologists are not always available. Introduction of new features and learning algorithms is the future scope of the present work.\\

*\textbf{Corresponding Author Note-} On behalf of all authors, the corresponding author states that there is no conflict of interest.
\bibliographystyle{model1-num-names}
\bibliography{paper}

\begin{thebibliography}{32}
\expandafter\ifx\csname natexlab\endcsname\relax\def\natexlab#1{#1}\fi
\providecommand{\bibinfo}[2]{#2}
\ifx\xfnm\relax \def\xfnm[#1]{\unskip,\space#1}\fi
\bibitem[{Parekh and Mittra(2011)}]{article3}
\bibinfo{author}{R.~Parekh}, \bibinfo{author}{A.~Mittra},
\newblock \bibinfo{title}{2011 : Automated detection of skin diseases using
  texture features},
\newblock \bibinfo{journal}{Int. Journal of Engineering Science and Technology}
  \bibinfo{volume}{3} (\bibinfo{year}{2011}) \bibinfo{pages}{4801--4808}.
\bibitem[{{Das} et~al.(2013){Das}, {Pal}, {Mazumder}, {Sarkar}, {Gangopadhyay},
  and {Nasipuri}}]{6686372}
\bibinfo{author}{N.~{Das}}, \bibinfo{author}{A.~{Pal}},
  \bibinfo{author}{S.~{Mazumder}}, \bibinfo{author}{S.~{Sarkar}},
  \bibinfo{author}{D.~{Gangopadhyay}}, \bibinfo{author}{M.~{Nasipuri}},
\newblock \bibinfo{title}{An svm based skin disease identification using local
  binary patterns},
\newblock in: \bibinfo{booktitle}{2013 Third International Conference on
  Advances in Computing and Communications}, pp. \bibinfo{pages}{208--211}.
\bibitem[{Pal et~al.(2013)Pal, Das, Sarkar, Gangopadhyay, and
  Nasipuri}]{10.1007/978-3-642-45062-4_48}
\bibinfo{author}{A.~Pal}, \bibinfo{author}{N.~Das},
  \bibinfo{author}{S.~Sarkar}, \bibinfo{author}{D.~Gangopadhyay},
  \bibinfo{author}{M.~Nasipuri},
\newblock \bibinfo{title}{A new rotation invariant weber local descriptor for
  recognition of skin diseases},
\newblock in: \bibinfo{editor}{P.~Maji}, \bibinfo{editor}{A.~Ghosh},
  \bibinfo{editor}{M.~N. Murty}, \bibinfo{editor}{K.~Ghosh},
  \bibinfo{editor}{S.~K. Pal} (Eds.), \bibinfo{booktitle}{Pattern Recognition
  and Machine Intelligence}, \bibinfo{publisher}{Springer Berlin Heidelberg},
  \bibinfo{address}{Berlin, Heidelberg}, \bibinfo{year}{2013}, pp.
  \bibinfo{pages}{355--360}.
\bibitem[{Wei et~al.(2018)Wei, Gan, and Ji}]{wei_gan_ji_2018}
\bibinfo{author}{L.-S. Wei}, \bibinfo{author}{Q.~Gan}, \bibinfo{author}{T.~Ji},
\newblock \bibinfo{title}{Skin disease recognition method based on image color
  and texture features},
\newblock \bibinfo{journal}{Computational and Mathematical Methods in Medicine}
  \bibinfo{volume}{2018} (\bibinfo{year}{2018}) \bibinfo{pages}{1–10}.
\bibitem[{{Kabari} and {Bakpo}(2009)}]{5409725}
\bibinfo{author}{L.~G. {Kabari}}, \bibinfo{author}{F.~S. {Bakpo}},
\newblock \bibinfo{title}{Diagnosing skin diseases using an artificial neural
  network},
\newblock in: \bibinfo{booktitle}{2009 2nd International Conference on Adaptive
  Science Technology (ICAST)}, pp. \bibinfo{pages}{187--191}.
\bibitem[{{Abdul-Rahman} et~al.(2012){Abdul-Rahman}, , {Yusoff}, {Mohamed}, and
  {Mutalib}}]{6498195}
\bibinfo{author}{S.~{Abdul-Rahman}}, , \bibinfo{author}{M.~{Yusoff}},
  \bibinfo{author}{A.~{Mohamed}}, \bibinfo{author}{S.~{Mutalib}},
\newblock \bibinfo{title}{Dermatology diagnosis with feature selection methods
  and artificial neural network},
\newblock in: \bibinfo{booktitle}{2012 IEEE-EMBS Conference on Biomedical
  Engineering and Sciences}, pp. \bibinfo{pages}{371--376}.
\bibitem[{{Yasir} et~al.(2014){Yasir}, {Rahman}, and {Ahmed}}]{7026918}
\bibinfo{author}{R.~{Yasir}}, \bibinfo{author}{M.~A. {Rahman}},
  \bibinfo{author}{N.~{Ahmed}},
\newblock \bibinfo{title}{Dermatological disease detection using image
  processing and artificial neural network},
\newblock in: \bibinfo{booktitle}{8th International Conference on Electrical
  and Computer Engineering}, pp. \bibinfo{pages}{687--690}.
\bibitem[{{Codella} et~al.(2017){Codella}, {Nguyen}, {Pankanti}, {Gutman},
  {Helba}, {Halpern}, and {Smith}}]{8030303}
\bibinfo{author}{N.~C.~F. {Codella}}, \bibinfo{author}{Q.~. {Nguyen}},
  \bibinfo{author}{S.~{Pankanti}}, \bibinfo{author}{D.~A. {Gutman}},
  \bibinfo{author}{B.~{Helba}}, \bibinfo{author}{A.~C. {Halpern}},
  \bibinfo{author}{J.~R. {Smith}},
\newblock \bibinfo{title}{Deep learning ensembles for melanoma recognition in
  dermoscopy images},
\newblock \bibinfo{journal}{IBM Journal of Research and Development}
  \bibinfo{volume}{61} (\bibinfo{year}{2017}) \bibinfo{pages}{5:1--5:15}.
\bibitem[{{Thao} and {Quang}(2017)}]{8233570}
\bibinfo{author}{L.~T. {Thao}}, \bibinfo{author}{N.~H. {Quang}},
\newblock \bibinfo{title}{Automatic skin lesion analysis towards melanoma
  detection},
\newblock in: \bibinfo{booktitle}{2017 21st Asia Pacific Symposium on
  Intelligent and Evolutionary Systems (IES)}, pp. \bibinfo{pages}{106--111}.
\bibitem[{Esteva et~al.(2017)Esteva, Kuprel, Novoa, Ko, M~Swetter, M~Blau, and
  Thrun}]{article}
\bibinfo{author}{A.~Esteva}, \bibinfo{author}{B.~Kuprel},
  \bibinfo{author}{R.~Novoa}, \bibinfo{author}{J.~Ko},
  \bibinfo{author}{S.~M~Swetter}, \bibinfo{author}{H.~M~Blau},
  \bibinfo{author}{S.~Thrun},
\newblock \bibinfo{title}{Dermatologist-level classification of skin cancer
  with deep neural networks},
\newblock \bibinfo{journal}{Nature} \bibinfo{volume}{542}
  (\bibinfo{year}{2017}).
\bibitem[{Li and Shen(2017)}]{article1}
\bibinfo{author}{Y.~Li}, \bibinfo{author}{L.~Shen},
\newblock \bibinfo{title}{Skin lesion analysis towards melanoma detection using
  deep learning network},
\newblock \bibinfo{journal}{Sensors} \bibinfo{volume}{18}
  (\bibinfo{year}{2017}).
\bibitem[{{Shamsul Arifin} et~al.(2012){Shamsul Arifin}, {Golam Kibria},
  {Firoze}, and and}]{6359626}
\bibinfo{author}{M.~{Shamsul Arifin}}, \bibinfo{author}{M.~{Golam Kibria}},
  \bibinfo{author}{A.~{Firoze}}, \bibinfo{author}{M.~A. and},
\newblock \bibinfo{title}{Dermatological disease diagnosis using color-skin
  images},
\newblock in: \bibinfo{booktitle}{2012 International Conference on Machine
  Learning and Cybernetics}, volume~\bibinfo{volume}{5}, pp.
  \bibinfo{pages}{1675--1680}.
\bibitem[{Delgado~Gomez et~al.(2007)Delgado~Gomez, Butakoff, Ersb{\o}ll, and
  Carstensen}]{DelgadoGomez:2007:ACD:1238135.1238180}
\bibinfo{author}{D.~Delgado~Gomez}, \bibinfo{author}{C.~Butakoff},
  \bibinfo{author}{B.~Ersb{\o}ll}, \bibinfo{author}{J.~M. Carstensen},
\newblock \bibinfo{title}{Automatic change detection and quantification of
  dermatological diseases with an application to psoriasis images},
\newblock \bibinfo{journal}{Pattern Recogn. Lett.} \bibinfo{volume}{28}
  (\bibinfo{year}{2007}) \bibinfo{pages}{1012--1018}.
\bibitem[{Kundu et~al.(2011)Kundu, Das, and
  Nasipuri}]{DBLP:journals/corr/abs-1103-0120}
\bibinfo{author}{S.~Kundu}, \bibinfo{author}{N.~Das},
  \bibinfo{author}{M.~Nasipuri},
\newblock \bibinfo{title}{Automatic detection of ringworm using local binary
  pattern {(LBP)}},
\newblock \bibinfo{journal}{CoRR} \bibinfo{volume}{abs/1103.0120}
  (\bibinfo{year}{2011}).
\bibitem[{Übeylı and İnan Güler(2005)}]{UBEYLI2005421}
\bibinfo{author}{E.~D. Übeylı}, \bibinfo{author}{İnan Güler},
\newblock \bibinfo{title}{Automatic detection of erythemato-squamous diseases
  using adaptive neuro-fuzzy inference systems},
\newblock \bibinfo{journal}{Computers in Biology and Medicine}
  \bibinfo{volume}{35} (\bibinfo{year}{2005}) \bibinfo{pages}{421 -- 433}.
\bibitem[{{Schwarz} et~al.(2017){Schwarz}, {Soliman}, {Omar}, {Buehler},
  {Ovsepian}, {Aguirre}, and {Ntziachristos}}]{7865979}
\bibinfo{author}{M.~{Schwarz}}, \bibinfo{author}{D.~{Soliman}},
  \bibinfo{author}{M.~{Omar}}, \bibinfo{author}{A.~{Buehler}},
  \bibinfo{author}{S.~V. {Ovsepian}}, \bibinfo{author}{J.~{Aguirre}},
  \bibinfo{author}{V.~{Ntziachristos}},
\newblock \bibinfo{title}{Optoacoustic dermoscopy of the human skin: Tuning
  excitation energy for optimal detection bandwidth with fast and deep
  imagingin vivo},
\newblock \bibinfo{journal}{IEEE Transactions on Medical Imaging}
  \bibinfo{volume}{36} (\bibinfo{year}{2017}) \bibinfo{pages}{1287--1296}.
\bibitem[{Chao et~al.(2017)Chao, K.~Meenan, and K.~Ferris}]{article2}
\bibinfo{author}{E.~Chao}, \bibinfo{author}{C.~K.~Meenan},
  \bibinfo{author}{L.~K.~Ferris},
\newblock \bibinfo{title}{Smartphone-based applications for skin monitoring and
  melanoma detection},
\newblock \bibinfo{journal}{Dermatologic Clinics} \bibinfo{volume}{35}
  (\bibinfo{year}{2017}).
\bibitem[{{Haralick} et~al.(1973){Haralick}, {Shanmugam}, and
  {Dinstein}}]{4309314}
\bibinfo{author}{R.~M. {Haralick}}, \bibinfo{author}{K.~{Shanmugam}},
  \bibinfo{author}{I.~{Dinstein}},
\newblock \bibinfo{title}{Textural features for image classification},
\newblock \bibinfo{journal}{IEEE Transactions on Systems, Man, and Cybernetics}
  \bibinfo{volume}{SMC-3} (\bibinfo{year}{1973}) \bibinfo{pages}{610--621}.
\bibitem[{{Conners} and {Harlow}(1976)}]{4045583}
\bibinfo{author}{R.~W. {Conners}}, \bibinfo{author}{C.~A. {Harlow}},
\newblock \bibinfo{title}{Some theoretical considerations concerning texture
  analysis of radiographic images},
\newblock in: \bibinfo{booktitle}{1976 IEEE Conference on Decision and Control
  including the 15th Symposium on Adaptive Processes}, pp.
  \bibinfo{pages}{162--167}.
\bibitem[{Assefa et~al.(2010)Assefa, Mansinha, Tiampo, Rasmussen, and
  Abdella}]{ASSEFA20101825}
\bibinfo{author}{D.~Assefa}, \bibinfo{author}{L.~Mansinha},
  \bibinfo{author}{K.~F. Tiampo}, \bibinfo{author}{H.~Rasmussen},
  \bibinfo{author}{K.~Abdella},
\newblock \bibinfo{title}{Local quaternion fourier transform and color image
  texture analysis},
\newblock \bibinfo{journal}{Signal Processing} \bibinfo{volume}{90}
  (\bibinfo{year}{2010}) \bibinfo{pages}{1825 -- 1835}.
\bibitem[{Jing et~al.(2006)Jing, Wong, and
  Zhang}]{Jing:2006:FRB:1161716.1161722}
\bibinfo{author}{X.-Y. Jing}, \bibinfo{author}{H.-S. Wong},
  \bibinfo{author}{D.~Zhang},
\newblock \bibinfo{title}{Face recognition based on discriminant fractional
  fourier feature extraction},
\newblock \bibinfo{journal}{Pattern Recogn. Lett.} \bibinfo{volume}{27}
  (\bibinfo{year}{2006}) \bibinfo{pages}{1465--1471}.
\bibitem[{{Ahonen} et~al.(2006){Ahonen}, {Hadid}, and {Pietikainen}}]{1717463}
\bibinfo{author}{T.~{Ahonen}}, \bibinfo{author}{A.~{Hadid}},
  \bibinfo{author}{M.~{Pietikainen}},
\newblock \bibinfo{title}{Face description with local binary patterns:
  Application to face recognition},
\newblock \bibinfo{journal}{IEEE Transactions on Pattern Analysis and Machine
  Intelligence} \bibinfo{volume}{28} (\bibinfo{year}{2006})
  \bibinfo{pages}{2037--2041}.
\bibitem[{{Caifeng Shan} et~al.(2005){Caifeng Shan}, {Shaogang Gong}, and
  {McOwan}}]{1530069}
\bibinfo{author}{{Caifeng Shan}}, \bibinfo{author}{{Shaogang Gong}},
  \bibinfo{author}{P.~W. {McOwan}},
\newblock \bibinfo{title}{Robust facial expression recognition using local
  binary patterns},
\newblock in: \bibinfo{booktitle}{IEEE International Conference on Image
  Processing 2005}, volume~\bibinfo{volume}{2}, pp. \bibinfo{pages}{II--370}.
\bibitem[{Jun et~al.(2011)Jun, Kim, and Kim}]{1890178}
\bibinfo{author}{B.~Jun}, \bibinfo{author}{T.~Kim}, \bibinfo{author}{D.~Kim},
\newblock \bibinfo{title}{A compact local binary pattern using maximization of
  mutual information for face analysis},
\newblock \bibinfo{journal}{Pattern Recogn.} \bibinfo{volume}{44}
  (\bibinfo{year}{2011}) \bibinfo{pages}{532--543}.
\bibitem[{Liao et~al.(2007)Liao, Zhu, Lei, Zhang, and Li}]{74549}
\bibinfo{author}{S.~Liao}, \bibinfo{author}{X.~Zhu}, \bibinfo{author}{Z.~Lei},
  \bibinfo{author}{L.~Zhang}, \bibinfo{author}{S.~Z. Li},
\newblock \bibinfo{title}{Learning multi-scale block local binary patterns for
  face recognition},
\newblock in: \bibinfo{editor}{S.-W. Lee}, \bibinfo{editor}{S.~Z. Li} (Eds.),
  \bibinfo{booktitle}{Advances in Biometrics}, \bibinfo{publisher}{Springer
  Berlin Heidelberg}, \bibinfo{address}{Berlin, Heidelberg},
  \bibinfo{year}{2007}, pp. \bibinfo{pages}{828--837}.
\bibitem[{Shan et~al.(2009)Shan, Gong, and McOwan}]{SHAN2009803}
\bibinfo{author}{C.~Shan}, \bibinfo{author}{S.~Gong}, \bibinfo{author}{P.~W.
  McOwan},
\newblock \bibinfo{title}{Facial expression recognition based on local binary
  patterns: A comprehensive study},
\newblock \bibinfo{journal}{Image and Vision Computing} \bibinfo{volume}{27}
  (\bibinfo{year}{2009}) \bibinfo{pages}{803 -- 816}.
\bibitem[{{Xu} et~al.(2010){Xu}, {Cha}, {Heyman}, {Venugopalan}, {Abiantun},
  and {Savvides}}]{5634504}
\bibinfo{author}{J.~{Xu}}, \bibinfo{author}{M.~{Cha}}, \bibinfo{author}{J.~L.
  {Heyman}}, \bibinfo{author}{S.~{Venugopalan}},
  \bibinfo{author}{R.~{Abiantun}}, \bibinfo{author}{M.~{Savvides}},
\newblock \bibinfo{title}{Robust local binary pattern feature sets for
  periocular biometric identification},
\newblock in: \bibinfo{booktitle}{2010 Fourth IEEE International Conference on
  Biometrics: Theory, Applications and Systems (BTAS)}, pp.
  \bibinfo{pages}{1--8}.
\bibitem[{Lian and Lu(2006)}]{11760023}
\bibinfo{author}{H.-C. Lian}, \bibinfo{author}{B.-L. Lu},
\newblock \bibinfo{title}{Multi-view gender classification using local binary
  patterns and support vector machines},
\newblock in: \bibinfo{editor}{J.~Wang}, \bibinfo{editor}{Z.~Yi},
  \bibinfo{editor}{J.~M. Zurada}, \bibinfo{editor}{B.-L. Lu},
  \bibinfo{editor}{H.~Yin} (Eds.), \bibinfo{booktitle}{Advances in Neural
  Networks - ISNN 2006}, \bibinfo{publisher}{Springer Berlin Heidelberg},
  \bibinfo{address}{Berlin, Heidelberg}, \bibinfo{year}{2006}, pp.
  \bibinfo{pages}{202--209}.
\bibitem[{Sun et~al.(2006)Sun, Zheng, Sun, Zou, and Zhao}]{11760023_29}
\bibinfo{author}{N.~Sun}, \bibinfo{author}{W.~Zheng}, \bibinfo{author}{C.~Sun},
  \bibinfo{author}{C.~Zou}, \bibinfo{author}{L.~Zhao},
\newblock \bibinfo{title}{Gender classification based on boosting local binary
  pattern},
\newblock in: \bibinfo{editor}{J.~Wang}, \bibinfo{editor}{Z.~Yi},
  \bibinfo{editor}{J.~M. Zurada}, \bibinfo{editor}{B.-L. Lu},
  \bibinfo{editor}{H.~Yin} (Eds.), \bibinfo{booktitle}{Advances in Neural
  Networks - ISNN 2006}, \bibinfo{publisher}{Springer Berlin Heidelberg},
  \bibinfo{address}{Berlin, Heidelberg}, \bibinfo{year}{2006}, pp.
  \bibinfo{pages}{194--201}.
\bibitem[{Nanni and Lumini(2008)}]{NANNI20083461}
\bibinfo{author}{L.~Nanni}, \bibinfo{author}{A.~Lumini},
\newblock \bibinfo{title}{Local binary patterns for a hybrid fingerprint
  matcher},
\newblock \bibinfo{journal}{Pattern Recognition} \bibinfo{volume}{41}
  (\bibinfo{year}{2008}) \bibinfo{pages}{3461 -- 3466}.
\bibitem[{{Hongliang Jin} et~al.(2004){Hongliang Jin}, {Qingshan Liu}, {Hanqing
  Lu}, and {Xiaofeng Tong}}]{1410446}
\bibinfo{author}{{Hongliang Jin}}, \bibinfo{author}{{Qingshan Liu}},
  \bibinfo{author}{{Hanqing Lu}}, \bibinfo{author}{{Xiaofeng Tong}},
\newblock \bibinfo{title}{Face detection using improved lbp under bayesian
  framework},
\newblock in: \bibinfo{booktitle}{Third International Conference on Image and
  Graphics (ICIG'04)}, pp. \bibinfo{pages}{306--309}.
\bibitem[{Kundu et~al.(2011)Kundu, Das, and Nasipuri}]{kundu_local}
\bibinfo{author}{S.~Kundu}, \bibinfo{author}{N.~Das},
  \bibinfo{author}{M.~Nasipuri},
\newblock \bibinfo{title}{Automatic detection of ringworm using local binary
  pattern (lbp)},
\newblock \bibinfo{journal}{Computing Research Repository - CORR}
  (\bibinfo{year}{2011}).

\end{thebibliography}

\end{document}